\documentclass[10pt,twocolumn,letterpaper]{article}

\usepackage{cvpr}
\usepackage{times}
\usepackage{epsfig}
\usepackage{graphicx}
\usepackage{amsmath}
\usepackage{amssymb}
\usepackage{subcaption}
\usepackage{color}
\usepackage{float}
\floatstyle{ruled}
\newfloat{algorithm}{tbp}{loa}
\providecommand{\algorithmname}{Algorithm}
\floatname{algorithm}{\protect\algorithmname}

\usepackage{algorithmic}

\usepackage[pagebackref=true,breaklinks=true,letterpaper=true,colorlinks,bookmarks=false]{hyperref}

\usepackage[latin9]{inputenc}

\providecommand{\algorithmname}{Algorithm}
\floatname{algorithm}{\protect\algorithmname}

 \cvprfinalcopy 

\def\cvprPaperID{2122} 

\ifcvprfinal\fi
\begin{document}

\title{Improving Human Activity Recognition Through Ranking and Re-ranking}

\author{Zhenzhong Lan, Shoou-I Yu, Alexander G. Hauptmann\\
Carnegie Mellon University\\
{\tt\small {lanzhzh, iyu, alex}@cs.cmu.edu}
}

\maketitle

\begin{abstract}
 We propose two well-motivated ranking-based methods to enhance the performance of current state-of-the-art human activity recognition systems. First, as an improvement over the classic power normalization method, we propose a parameter-free ranking technique called rank normalization (RaN). RaN normalizes each dimension of the video features to address the sparse and bursty distribution problems of Fisher Vectors and VLAD. Second, inspired by curriculum learning, we introduce a training-free re-ranking technique called multi-class iterative re-ranking (MIR). MIR captures relationships among action classes by separating easy and typical videos from difficult ones and re-ranking the prediction scores of classifiers accordingly. We demonstrate that our methods significantly improve the performance of state-of-the-art motion features on six real-world datasets.   
\end{abstract}

\section{Introduction}

It is a challenging task to recognize human activities among videos, especially unconstrained Internet videos with large visual diversity. A typical human activity recognition pipeline is often composed of the following three steps: (i) extract local video descriptors (e.g., STIP \cite{laptev2005space}, IDT \cite{wang2013action} or Deep learning features \cite{xu2014discriminative}), (ii) encode and pool the local descriptors into video descriptors (e.g., Fisher Vectors \cite{perronnin2010improving} or VLADs \cite{arandjelovic2013all}), and (iii) classify the video representations (e.g., SVM). The basic components of the current state-of-the-art human activity recognition system, for example, are Improved Dense Trajectories (IDT), Fisher Vectors, and SVMs. In this paper, we introduce two ranking-based methods to improve the performance of this state-of-the-art human activity recognition pipeline.

The first proposed method is to tackle the sparse \cite{perronnin2010improving} and bursty \cite{arandjelovic2013all} distribution problems of Fisher Vectors \cite{perronnin2010improving} and VLADs \cite{arandjelovic2013all}. A common issue of Fisher Vector and VLAD encoding is that they generate high dimensional sparse and bursty video representations whose similarity cannot be accurately assessed by linear measurements. The sparse distribution, i.e. most of the values in the representation are close to zero, comes from the fact that as the codebook size increases, fewer local descriptors are assigned to each codebook. A bursty distribution, i.e. the values of a few dimensions dominate the representation, occurs when repeated patterns in a video generate a few artificially large components in the representation. To deal with those problems, Perronnin \textit{et al.}\cite{perronnin2010improving} invented a simple Power Normalization (PN) method to disperse the video representations. PN uses an element-wise power operation to discount large values and increase small values of video representations. As one of the most significant improvements in the past few years, this simple algorithm essentially makes Fisher Vectors and VLADs useful in practice, and has been widely adopted by the research community to both handcrafted \cite{laptev2005space,wang2013action} and deeply-learned features \cite{xu2014discriminative, zha2015exploiting}. However, PN can only alleviate the sparse and bursty distribution problems. It is often difficult to decide how much dispersal each task requires. There is also no theoretical justification for using sign square root \cite{perronnin2010improving,arandjelovic2013all} as a rule of thumb. Such being the case, we propose Rank Normalization (RaN) that is parameter-free and thoroughly addresses the sparse bursty distribution problem. RaN ranks all video representations in a dataset along each dimension and uses the normalized rankings in place of the original video representations for the subsequent classification. We show that RaN also works for scenarios where not all data are available. Using an approximation method which requires only a subset of video representations gives a similar ranking performance to the precise case.

The second method we suggest is inspired by curriculum learning which mimics the human learning scheme and has become popular for image and video classification \cite{bengio2009curriculum, chen2015webly, jiang2015self, cimpoi2015deep}. Curriculum learning \cite{bengio2009curriculum,kumar2010self} suggests distinguishing easy and typical samples from difficult ones and treating them separately. Traditional curriculum learning methods  \cite{bengio2009curriculum,kumar2010self,bengio2009learning} rely on human or extra resources to define data with difficulty levels (curriculum). Instead, we use freely-available classifiers from other action classes to define the curriculum. We then re-rank the prediction results to promote easy videos and suppress difficult ones. Our proposed method, called Multi-class Iterative Re-ranking (MIR), is easy to implement and training-free. 

In the remainder of this paper, we review some relevant works about the improvement of feature encoding, mostly associated with Fisher Vectors and VLADs. We also briefly introduce recent studies on capturing relationships among multiple action classes and curriculum learning. We then describe our two proposed methods in details and demonstrate their performance gain over the baseline approach (composed of IDT, Fisher Vectors, and SVMs) for the action recognition task represented by Hollywood2 and Olympic Sports datasets. Next, we combine those two methods and report results accordingly on both action recognition and event detection \cite{lan2013cmu} tasks. A conclusion and discussion of future works follow in the end.

\section{Related Work}

Features and encoding methods are the major sources of breakthroughs in conventional video representations. Among them the trajectory based approaches \cite{wang2013action,jiang2012trajectory}, especially the Dense Trajectory (DT), IDT \cite{wang2011action, wang2013action}, and Fisher Vectors encoding, are the basis of current state-of-the-art algorithms. 

Fisher Vectors and VLADs are very similar encoding methods \cite{arandjelovic2013all} and both have been popular for image and video classification \cite{wang2013action,perronnin2010improving,arandjelovic2013all, xu2014discriminative, mironicua2015fisher}. In the original scheme \cite{perronnin2007fisher, jegou2012negative} they either do not require post processing \cite{perronnin2007fisher} or use only L2 normalization \cite{jegou2012negative}. Although L2 normalization can reduce the influence of background information and transform the linear kernel into an L2 similarity measurement \cite{perronnin2010improving}, it does not disperse the data. As a result, the original Fisher Vector encoding method showed inconclusive results compared to other state-of-the-art encoding methods \cite{perronnin2007fisher}. It is the introduction of PN \cite{perronnin2010improving, arandjelovic2013all} that significantly improved the performance of those encoding methods and thus made them useful in practice. PN alleviates the problem of sparse and bursty distribution of Fisher Vector and VLAD. Later on, as a special design for VLAD, Arandjelovic and Zisserman \cite{arandjelovic2013all} proposed Intra-normalization to further reduce the bursty distribution problem of VLADs. J{\'e}gou and Chum \cite{jegou2012negative} used PCA to decorrelate a low dimensional representation and adopted multiple clustering to reduce the quantization errors for VLADs. Nonetheless, those methods only alleviated the burstiness problem. The sparsity of the encoded descriptors still depends on that of the original data. Unlike the above two approaches, RaN normalizes each dimension of Fisher Vectors to a distribution close to the uniform distribution, regardless of how sparse the original data are.

Given the encoded features, state-of-the-art methods often use `one versus rest' SVM, which does not consider the relationships among action classes. To model those relationships, Bergamo \& Torresani \cite{bergamo2012meta} suggested a meta-class method for identifying related image classes based on misclassification errors from a validation set. Hou et al. \cite{hou2014damn} identified similar class pairs and grouped them together to train `two versus rest' classifiers. By combining `two versus rest' with `one versus rest' classifiers, they observed significant improvements from baselines. Unlike the aforementioned approaches that require training and modify the predictions for one time only, MIR is training-free and iteratively updates the prediction rankings given those from previous iterations. 

Our MIR model is inspired by a new learning paradigm called curriculum learning, proposed by Bengio et al. \cite{bengio2009curriculum}. Curriculum learning rates the difficulty levels for classifying samples and uses the rating as a `curriculum' to guide learning. This new way of learning, from a human behavioral perspective, is considered similar to human learning in principle \cite{khan2011humans}. Moreover, just like school curriculum design in the everyday case, it relies on human or other data resources to define the curriculum. Our method instead uses freely available classifier predictions from other action classes to help define the curriculum without human intervention.

\section{Action Recognition Preliminaries}

\label{action_recognition}

\paragraph{Problem formulation}

Given a collection of short clips of videos that usually last a few seconds, the goal of an action recognition task is to classify them into actions of interest such as \textit{running} and \textit{kissing}, solely based on the video content. By evaluating on this task, we dig deep into the characteristics of RaN and MIR. 

\paragraph{Benchmark datasets} We rely on two widely used action recognition benchmark datasets including the Hollywood2 \cite{marszalek2009actions} and Olympic Sports datasets \cite{niebles2010modeling}. 

The Hollywood2 dataset \cite{marszalek2009actions} contains 12 action classes and 1707 video clips that are collected from 69 different Hollywood movies.
There are 12 action classes such as answering a phone, driving a car and standing up. Each video of this dataset may contain multiple actions. We use the clean training dataset and standard splits with training  (823 samples) and test videos (884 samples) provided \cite{marszalek2009actions}. The performance is evaluated by
computing the average precision (AP) for each of the action classes and reporting the mean AP (mAP) over all classes . 
 
The Olympic Sports dataset \cite{niebles2010modeling} consists of 16 athletes practicing sports such as high-jump, pole-vault and basketball lay-up. It has a total of 783 video clips. We use standard splits with 649 training clips and 134 test clips and report mAP as in \cite{niebles2010modeling} for comparison purposes.

\paragraph{Experimental settings}
We follow the experimental settings in \cite{wang2013action}. More specifically, we use IDT features extracted using 15 frame tracking and camera motion stabilization. PCA is utilized to reduce the dimensionality of IDT descriptors by a factor of two. After reduction, the local descriptors are augmented with three-dimensional normalized location information \cite{lan2014beyond}. Fisher Vector encoding maps the raw descriptors into a Gaussian Mixture Model with 256 Gaussians trained from a set of 256000 randomly sampled data points. Classification in conducted by a `one versus rest' linear SVM classifier with a fixed $C=100$ (\cite{wang2013action}). For PN, we use sign square root unless otherwise stated.

\section{Rank Normalization (RaN)}

As discussed in \cite{perronnin2010improving,arandjelovic2013all}, Fisher Vector encoding often generates high dimensional sparse and bursty video representations, whose similarity cannot be accurately quantified by a linear kernel or an L2 similarity measurement. To address this problem, Perronnin \textit{et al.}\cite{perronnin2010improving} introduced PN, which has the following element-wise operation:
\[ f(z) = sign(z)|z|^{\alpha}, \] 
where $ 0 \leq \alpha \leq 1$.   PN can only alleviate the sparse and bursty distribution problems and it is difficult to determine a good $\alpha$ for different tasks. To overcome this weakness, we propose RaN, which is parameter-free and handles the bursty distributions in a more fundamental way. RaN applies to each dimension of the Fisher Vectors the following function: 
\[f(z) = rank(z)/N,\]
where rank(z) is z's position after sorting along the dimension of all $N$ Fisher Vectors in the dataset. After RaN, the values in each dimension of Fisher Vectors are spread out and have a distribution close to uniform.

\begin{figure}
\centering
    \begin{subfigure}[b]{0.23\textwidth}
                \includegraphics[width=\textwidth]{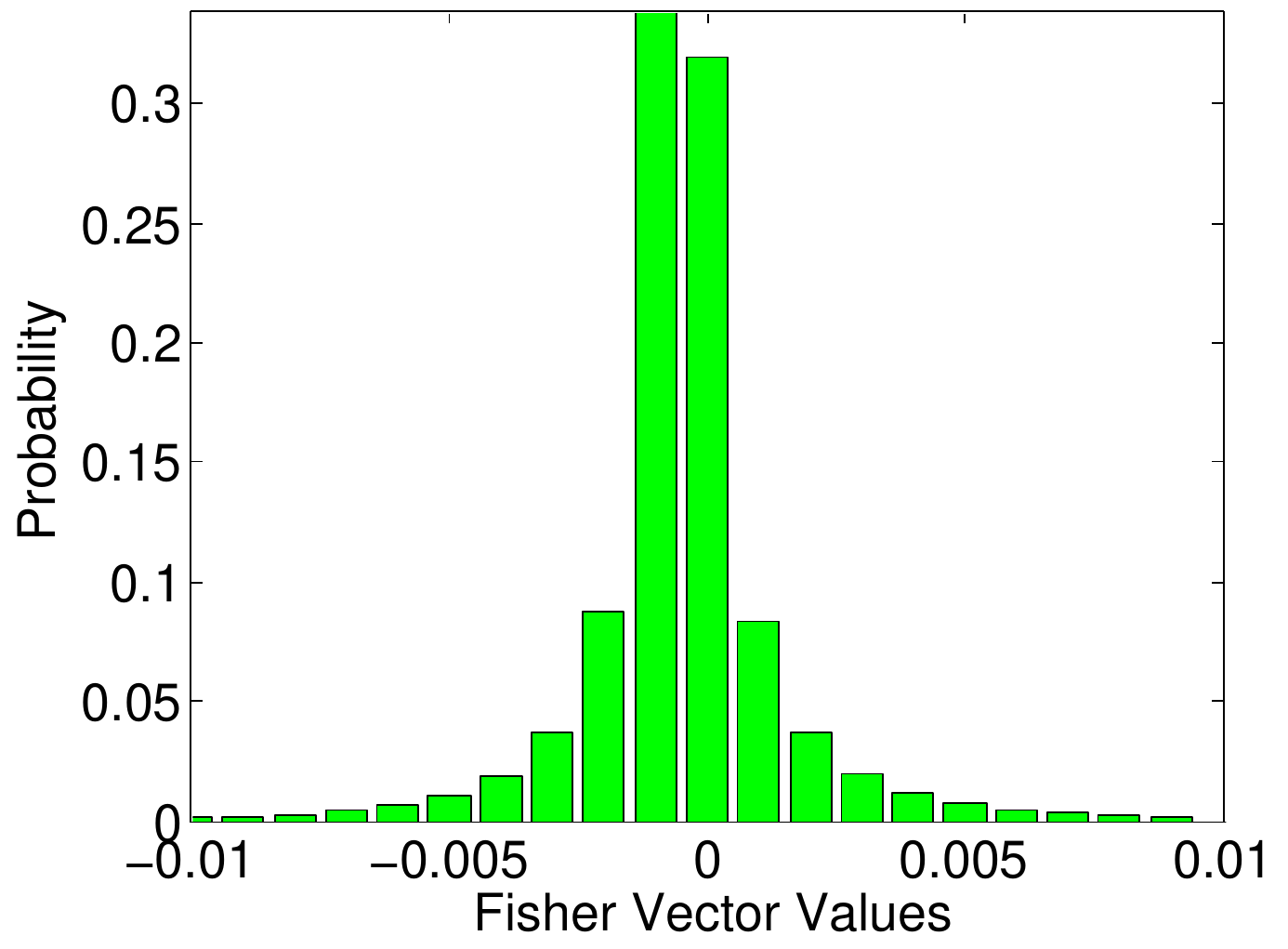}
                    \caption{L2 Normalization}
                    \label{fig:fa}
    \end{subfigure}
    \begin{subfigure}[b]{0.23\textwidth}
                \includegraphics[width=\textwidth]{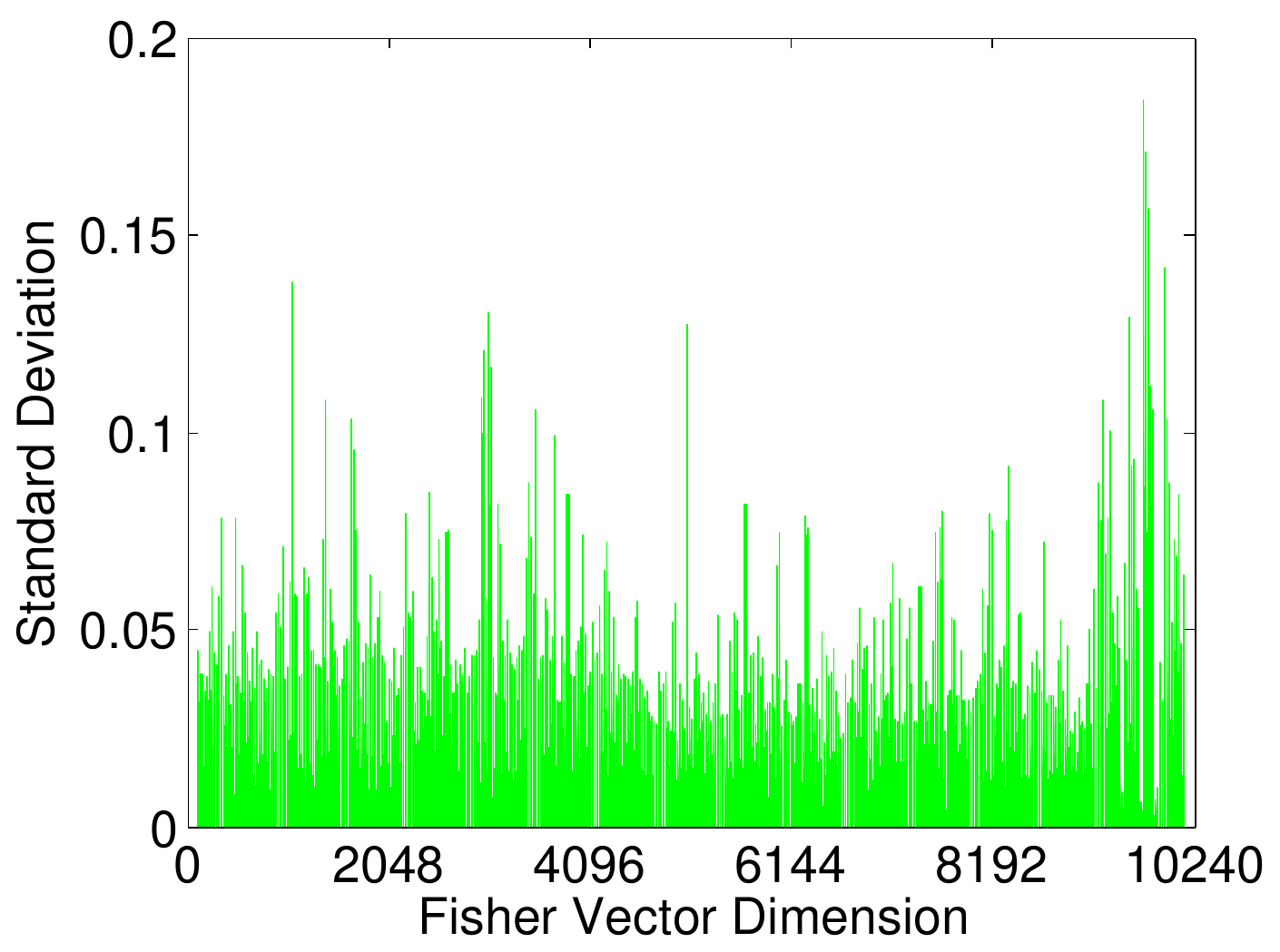}
                    \caption{L2 Normalization}
                    \label{fig:fb}
    \end{subfigure}
     \begin{subfigure}[b]{0.23\textwidth}
                \includegraphics[width=\textwidth]{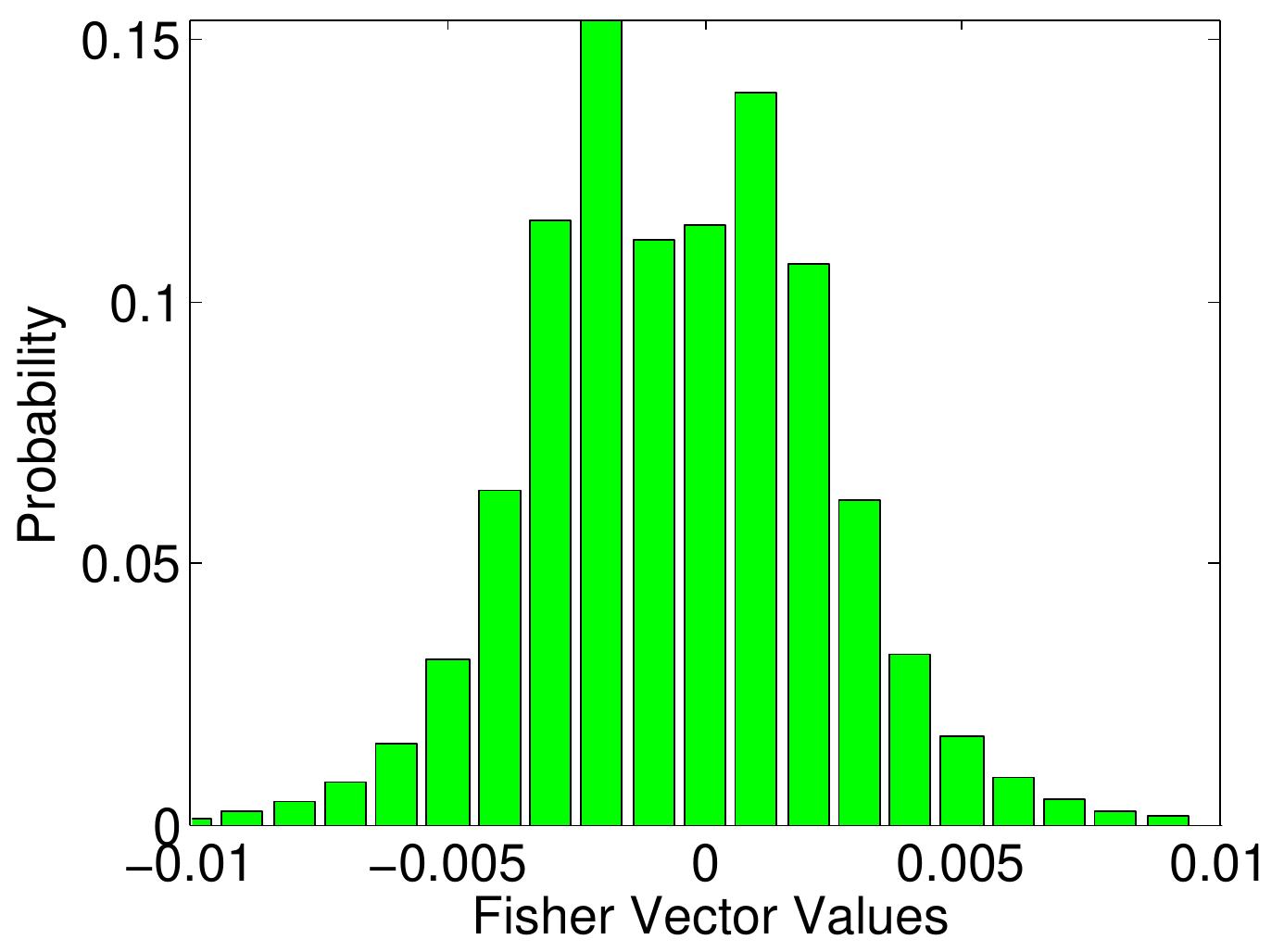}
                    \caption{PN}
                    \label{fig:fc}
    \end{subfigure}
        \begin{subfigure}[b]{0.23\textwidth}
                \includegraphics[width=\textwidth]{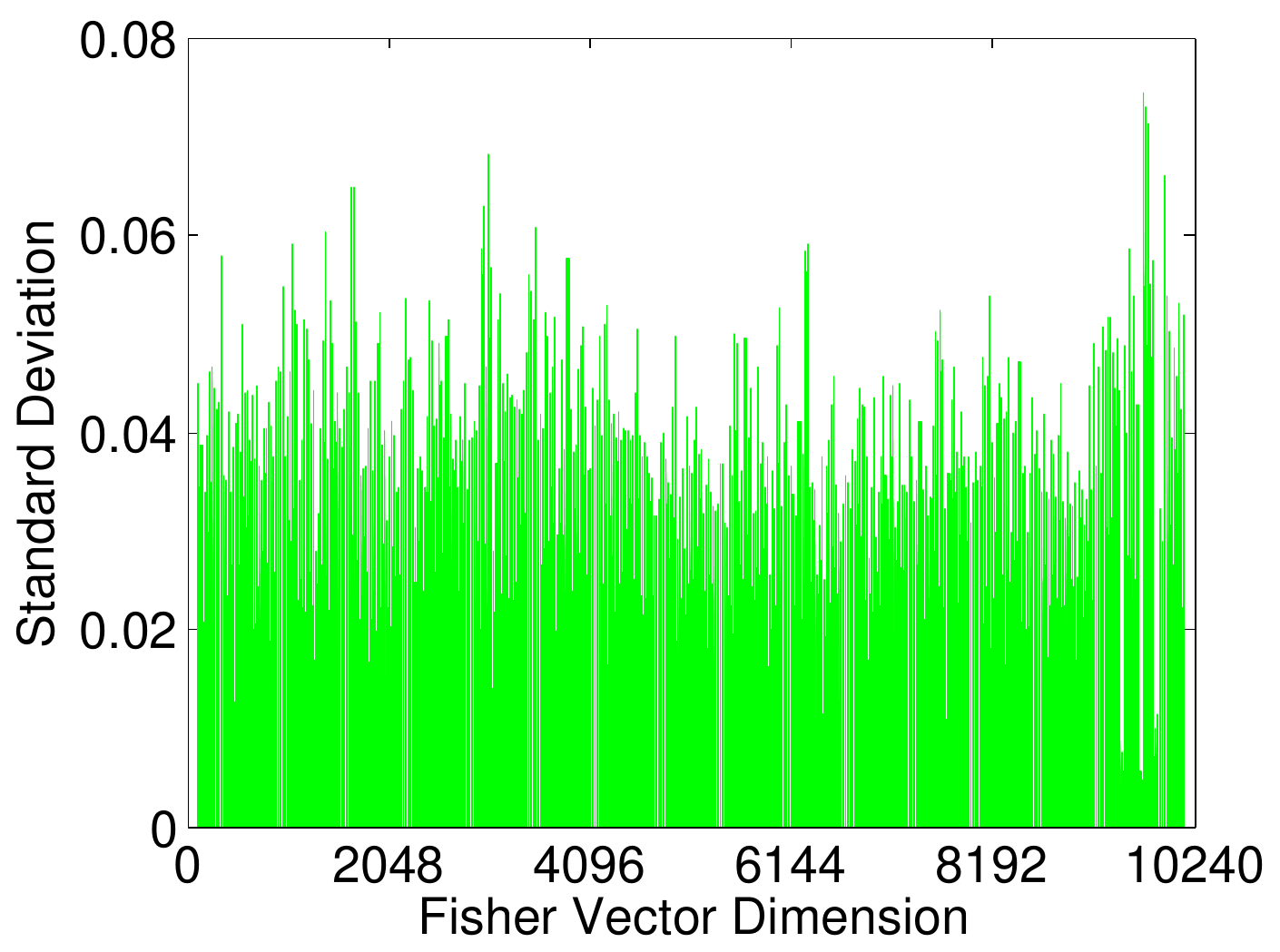}
                    \caption{PN}
                    \label{fig:fd}
    \end{subfigure}
    \begin{subfigure}[b]{0.23\textwidth}
                \includegraphics[width=\textwidth]{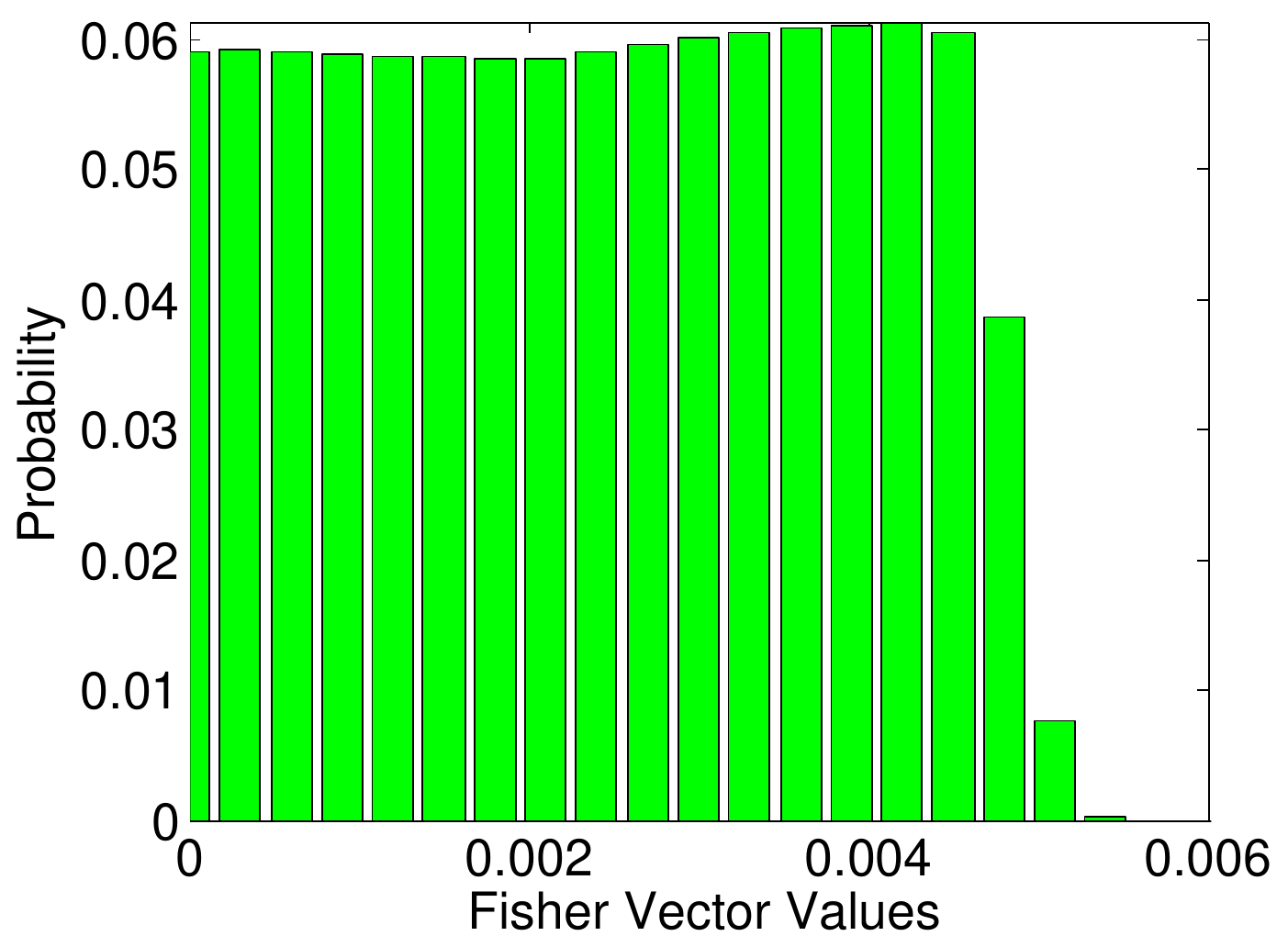}
                    \caption{RaN}
                     \label{fig:fe}
    \end{subfigure}
     \begin{subfigure}[b]{0.23\textwidth}
                \includegraphics[width=\textwidth]{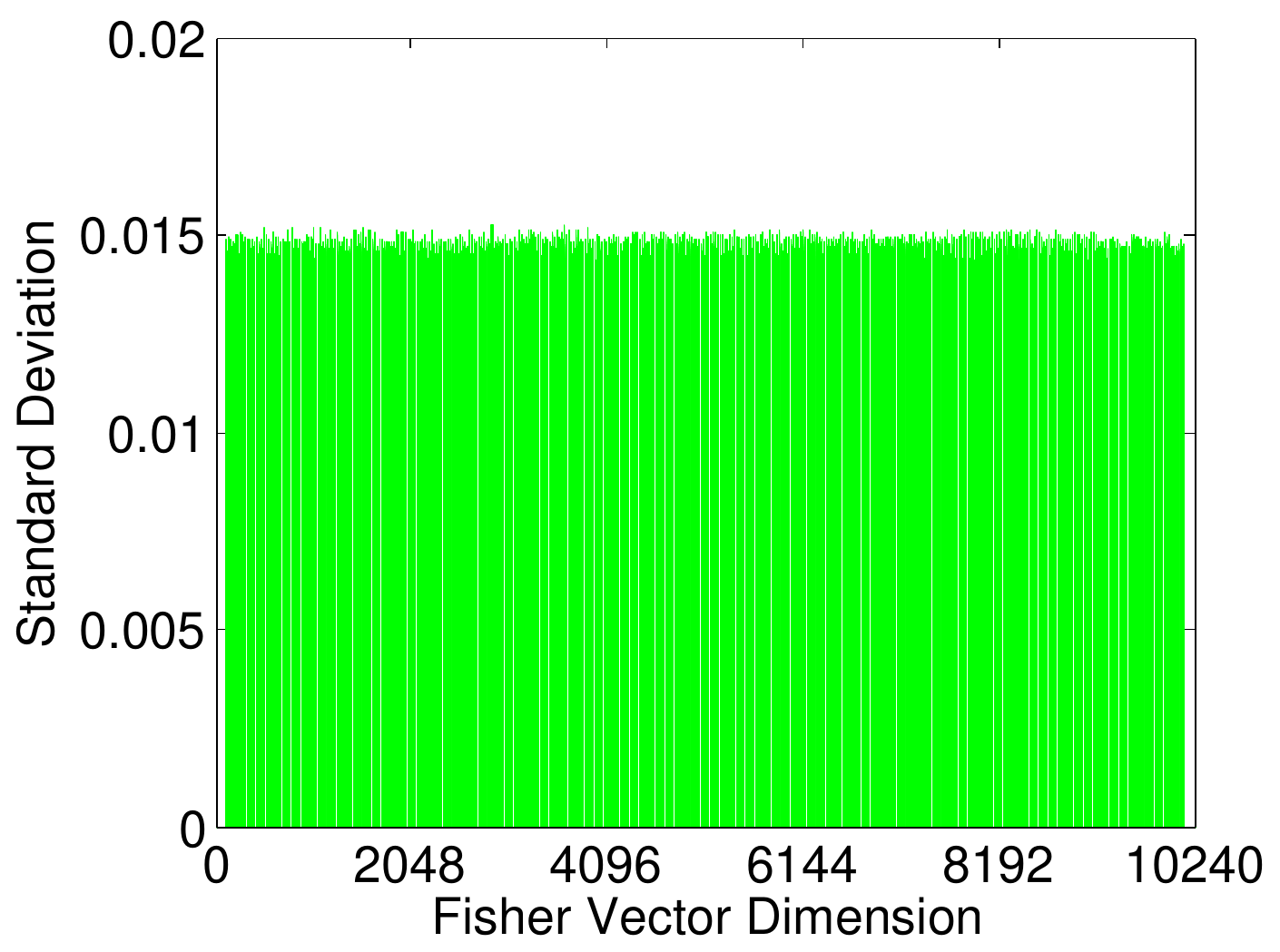}
                    \caption{RaN}
                    \label{fig:ff}
    \end{subfigure}
\caption{\textbf{The effect of different normalization methods on the sparse problem and the bursty problem of Fisher Vectors.} Plots on the left (\ref{fig:fa}, \ref{fig:fc}, \ref{fig:fe}) are the distributions of all the values in the Fisher Vectors across all videos for Hollywood2.  On the right (\ref{fig:fb}, \ref{fig:fd}, \ref{fig:ff}) are the standard deviations of the values in each dimension. L2 normalization serves as a baseline method and is applied after PN and RaN.}
\label{fig:rn}
\end{figure}

\begin{figure*}
\centering
    \begin{subfigure}[b]{0.33\textwidth}
                \includegraphics[width=\textwidth]{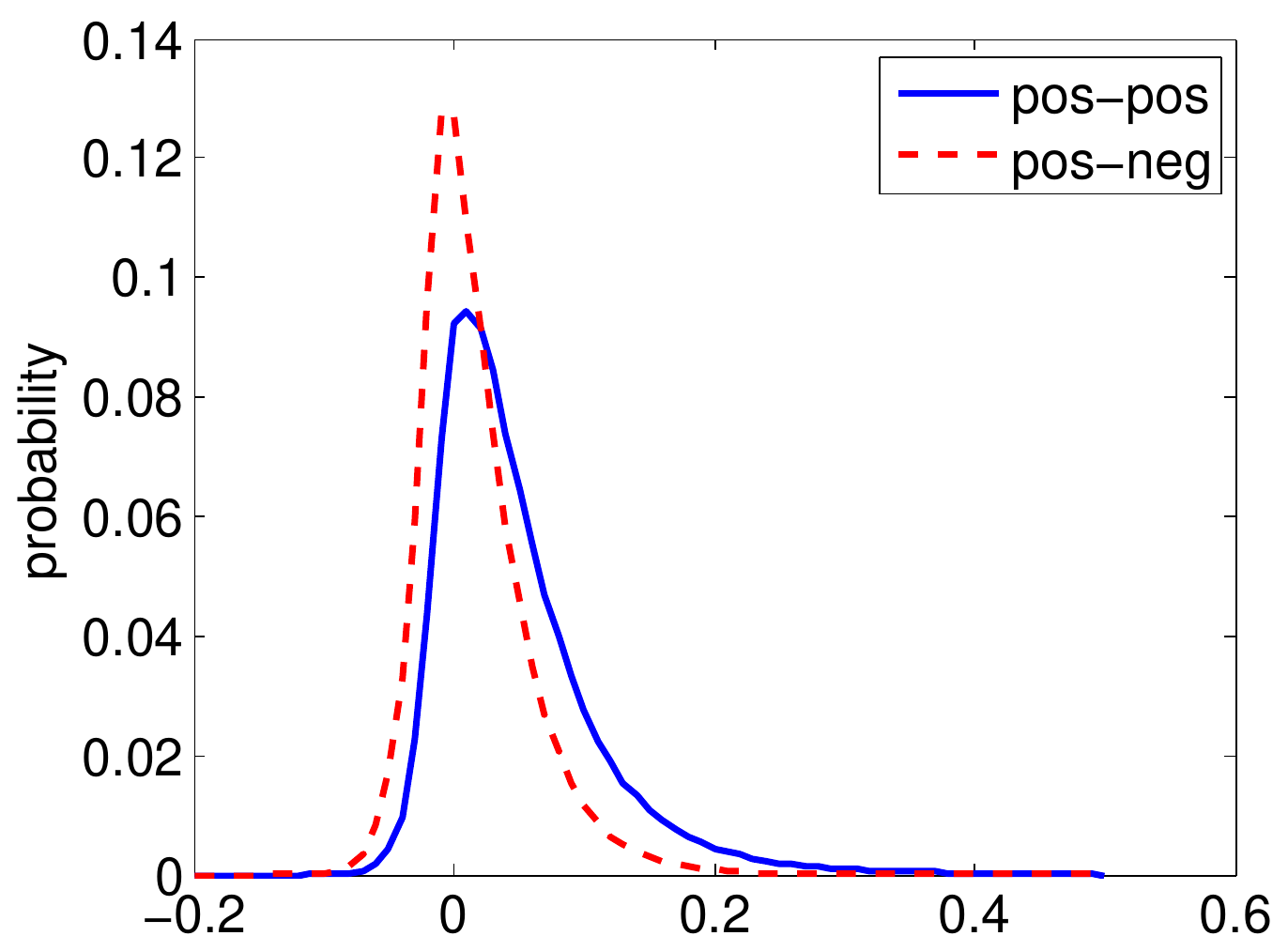}
                    \caption{L2 Normalization}
                    \label{fig:ca}
    \end{subfigure}
    \begin{subfigure}[b]{0.33\textwidth}
                \includegraphics[width=\textwidth]{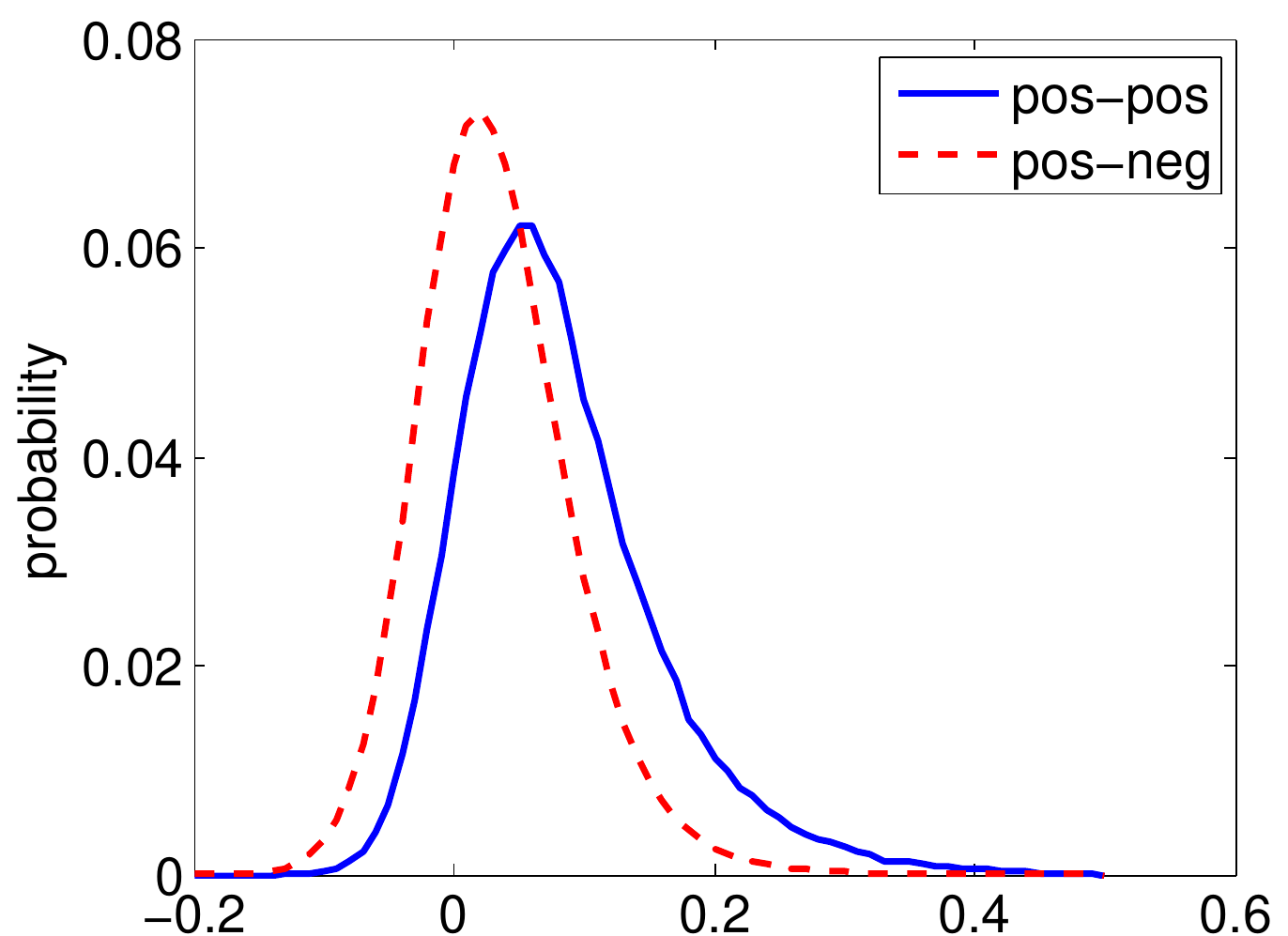}
                    \caption{PN}
                    \label{fig:cb}
    \end{subfigure}
     \begin{subfigure}[b]{0.33\textwidth}
                \includegraphics[width=\textwidth]{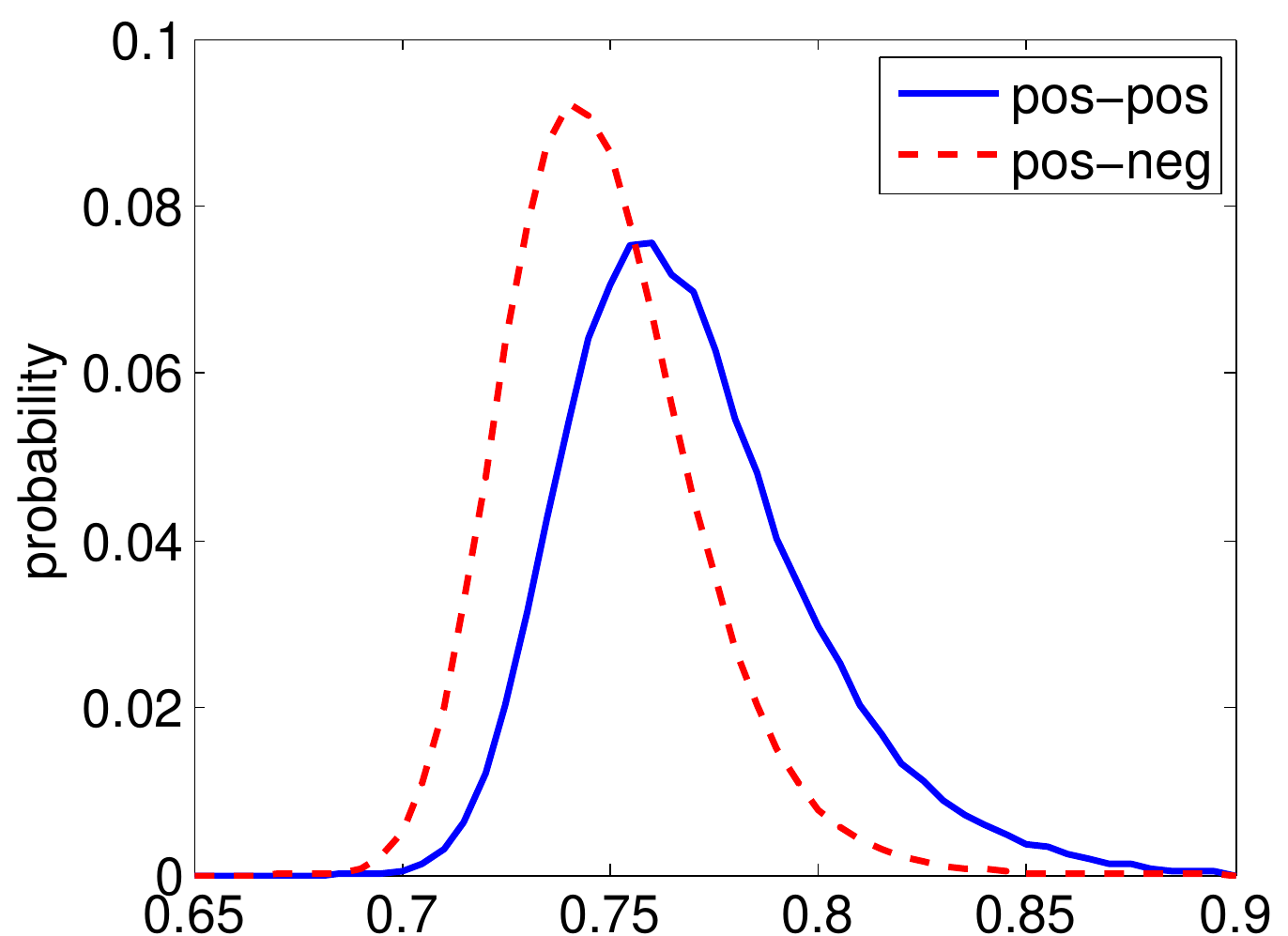}
                    \caption{RaN}
                    \label{fig:cc}
    \end{subfigure}
\caption{\textbf{The cosine similarities of videos under different normalization methods.} The blue and plain lines are the distributions of cosine similarities between positive samples, and the red and dashed lines show the distributions of cosine similarities between positive and negative samples.}
\label{fig:cosine} 
\end{figure*}

\subsection{Qualitative Analysis}

\begin{figure}
\centering
\includegraphics[width=0.47\textwidth]{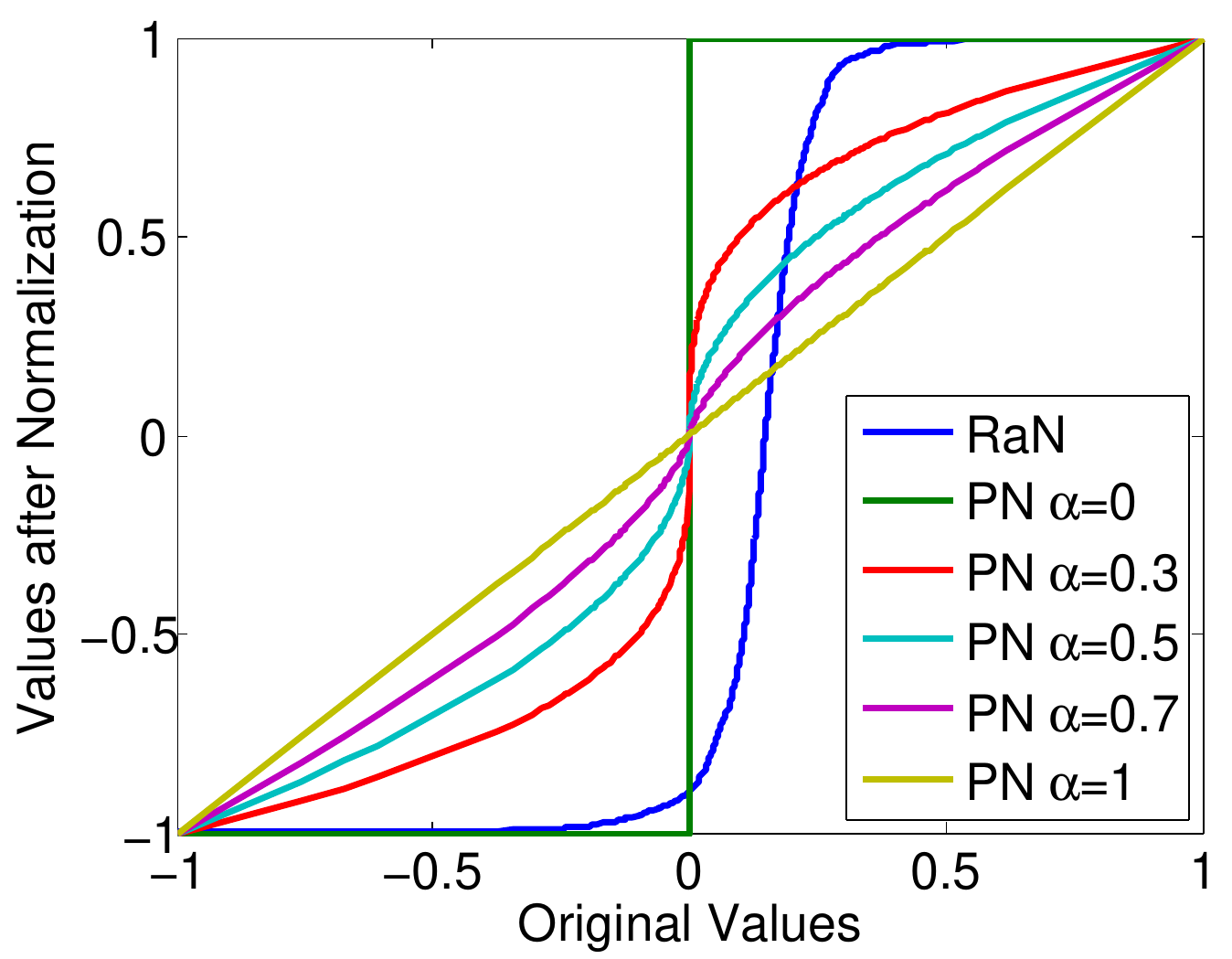}
\caption{ \textbf{Comparison of the effects of PN and RaN on the first dimension of Fisher Vectors in Hollywood2.} For a better visualization, we normalize all the curves so that their values are between -1 and 1. }
\label{fig:transform}
\end{figure}

\begin{figure}
\centering
    \begin{subfigure}[b]{0.23\textwidth}
                \includegraphics[width=\textwidth]{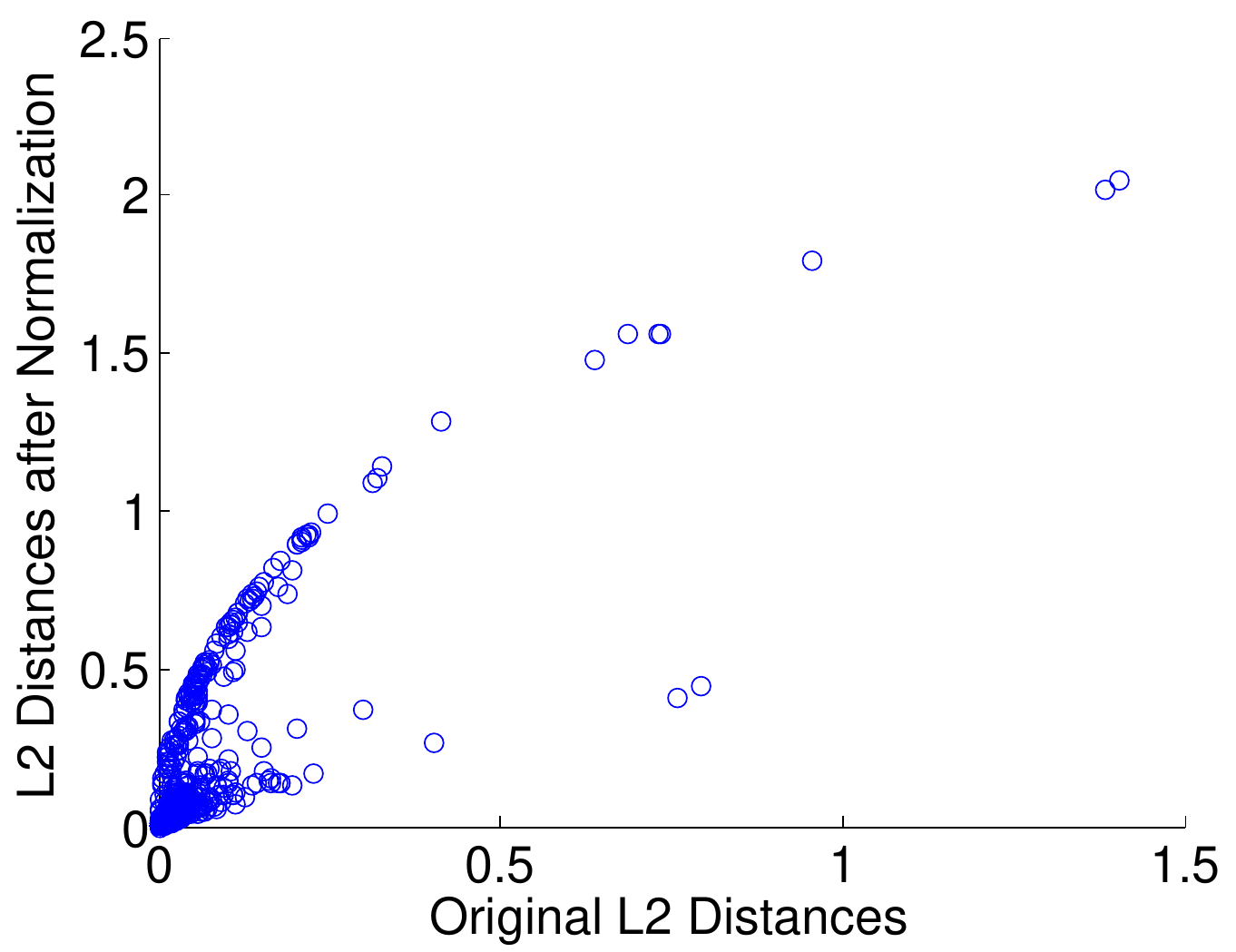}
                    \caption{PN}
                    \label{fig:fbbc}
    \end{subfigure}
        \begin{subfigure}[b]{0.23\textwidth}
                \includegraphics[width=\textwidth]{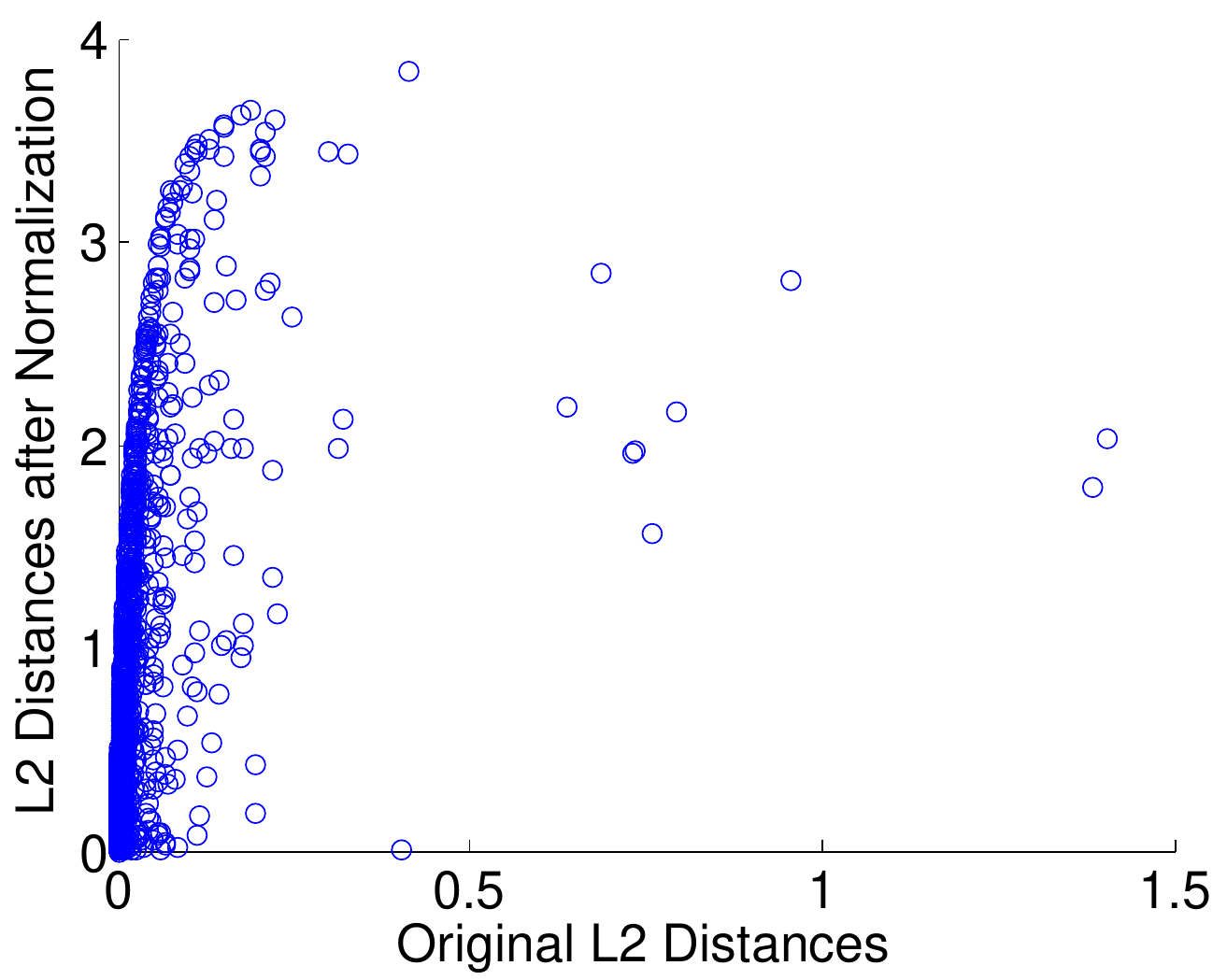}
                    \caption{RaN}
                    \label{fig:fabc}
    \end{subfigure}
\caption{\textbf{Comparison of L2 distances before and after normalization on the first dimension of Fisher Vectors in Hollywood2.} }
\label{fig:drn}
\end{figure}

To qualitatively characterize the differences between PN and RaN, we visualize their results from different perspectives using all Fisher Vectors in the Hollywood2 dataset. Results are shown in Figure \ref{fig:rn}, \ref{fig:cosine}, \ref{fig:transform}, and \ref{fig:drn}. Figure \ref{fig:rn} displays the distributions of Fisher Vector values and their standard deviations along each dimension. As evidenced by Figure \ref{fig:fa} and \ref{fig:fb}, the distribution of L2-normalized \textbf{Fisher Vector values are indeed sparse and bursty}. These sparse and bursty distributed representations would have cosine similarities that clustered around zero (Figure \ref{fig:ca}). PN disperses the data and reduces the probability of zero cosine similarities (Figure \ref{fig:cb}). However, the distributions after PN are still somewhat sparse (Figure \ref{fig:fc}) and bursty (Figure \ref{fig:fd}), whose cosine similarities remain centred around zero (Figure \ref{fig:cb}). Conversely,  RaN disperses the data to an extreme (Figure \ref{fig:fe}, \ref{fig:ff})  and completely removes zero cosine similarity (Figure \ref{fig:cc}).

In Figure \ref{fig:transform}, we also compare the normalization effects on the first dimension of Fisher Vectors on the Hollywood2 dataset. The x-axis shows the original values and the y-axis are the corresponding values after normalization. For a better visualization, we normalize all the curves so that the values are between -1 and 1. Basically, PN returns the original values when $\alpha=1$ and becomes a step function as $\alpha=0$. When $0<\alpha<1$ , PN carries out a mapping like a sigmoid activation function. It magnifies those values around zeros and allows them to possess more of the y-axis space. Through this magnification, PN assumes that values around zeros are more important than what their original values indicate. However, this assumption provides no information about how much magnification is sufficient. In contrast, RaN states that all Fisher Vector values are equally important in calculating the similarities between video representations.

Another interesting observation is that PN cannot dynamically adjust its mapping according to the distribution of values in different sign spaces whereas RaN can. We see from Figure \ref{fig:transform} that only a small subset of data are negative; while in the PN case, data with negative values possess as much space as data with positive values. Through dynamic mapping, RaN does much better than PN in spreading out the data (Figure \ref{fig:drn}). The x-axis of Figure \ref{fig:drn} shows the original distribution of L2 distances of the first dimension of Fisher Vectors and the y-axis shows the corresponding distances after normalization. Evidently, after PN, most of the L2 distances are still cluttered around zero; while after RaN, the distribution of L2 distances spread out significantly and become much more separable.

\subsection{Quantitative Analysis}

\begin{figure}
\centering
    \begin{subfigure}[b]{0.47\textwidth}
                \includegraphics[width=\textwidth]{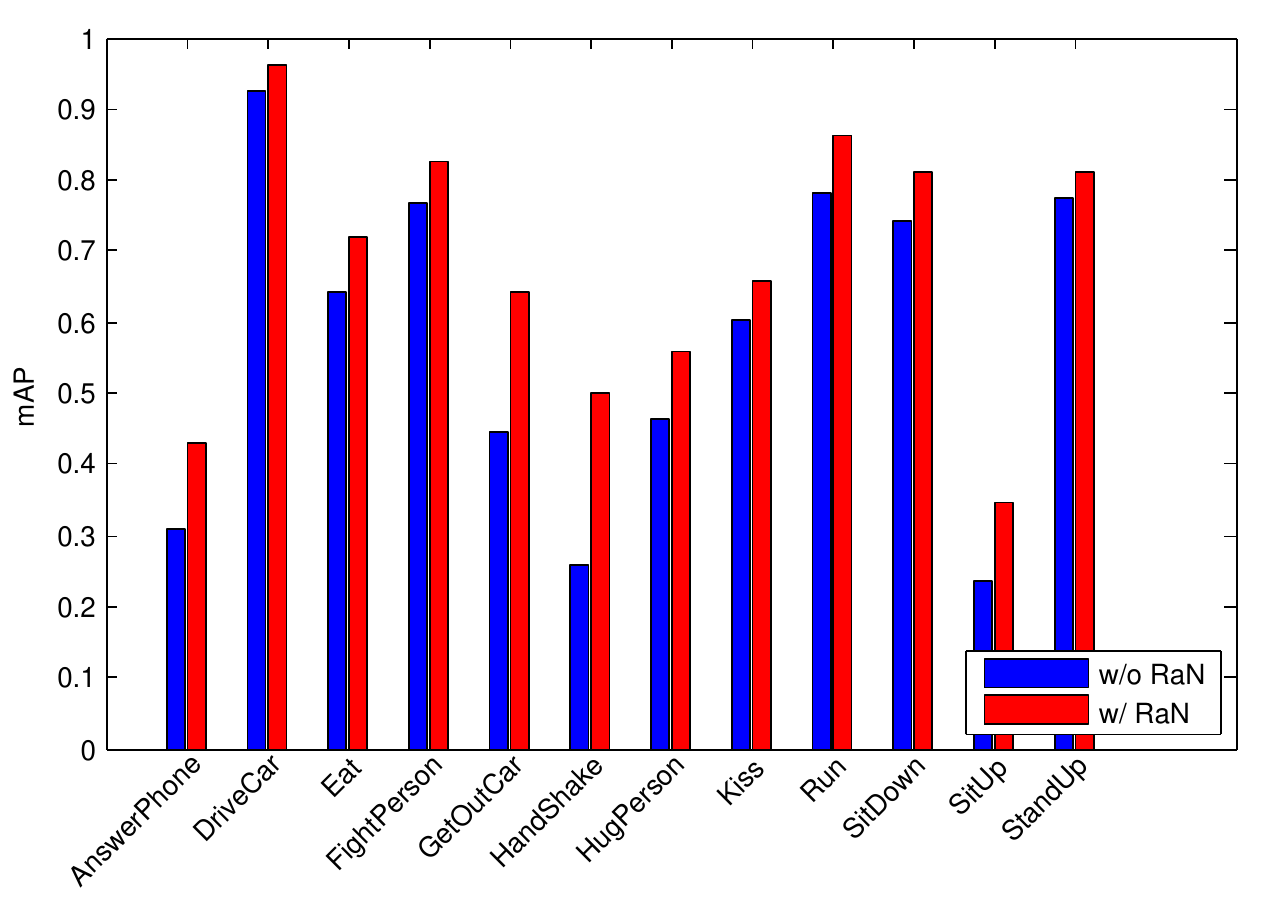}
                    \caption{Hollywood2}
                    \label{fig:ma}
    \end{subfigure}
    \begin{subfigure}[b]{0.47\textwidth}
                \includegraphics[width=\textwidth]{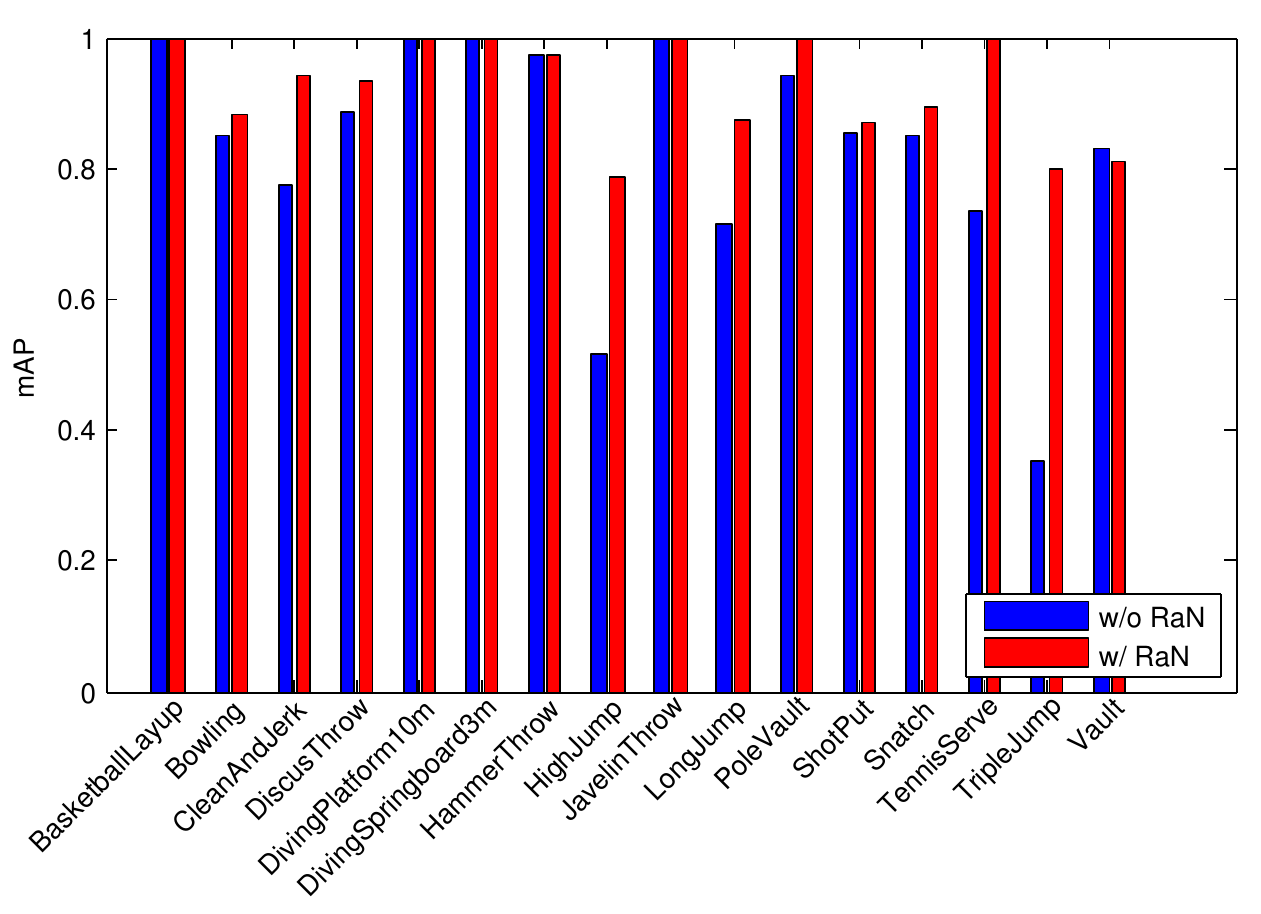}
                    \caption{Olympic Sports}
                    \label{fig:mb}
    \end{subfigure}
\caption{\textbf{Per-class performance comparison of the baseline performances with and without RaN.} }
\label{fig:ran_individual}
\end{figure}

In Table \ref{tab:rn}, we compare the results of different normalization methods on both Hollywood2 and Olympic Sports datasets. As in Figure \ref{fig:rn}, L2 normalization serves as a baseline method and is applied after PN and RaN. First, we show that PN improves the performance of the baseline method on both datasets by more than $6\%$. These improvements are significant given the difficulty level of the task. RaN further improves the PN performance by more than $1\%$ on Hollywood2 and $3\%$ on Olympic Sports. Compared to the baseline method, RaN gives about $10\%$ absolute improvement on a difficult dataset like Hollywood2; on the relatively easy one like Olympic Sports that has less potential to explore, RaN still manages to improve by more than $9\%$, absolutely. We also find that applying RaN to local descriptors within each video and combining the rankings with the original descriptors (Two-level RaN) can further improve the performance of Hollywood2 by about $1\%$, reaching an overall improvement of around $11\%$. We hypothesize that this is related to removing background distribution as suggested in \cite{zhang2009efficient}, but here we do not further investigate this connection. Figure \ref{fig:ran_individual} shows the per-class comparison of with and without RaN. Most remarkably, RaN improves the baseline method on all 12 actions from the Hollywood2 dataset. On some of the hard classes like `HandShake', RaN improves the baseline results by more than $15\%$. A similar trend can be seen in the Olympic Sports dataset. Compared to the baseline method, RaN either improves or delivers similar results on 15 out of 16 actions. These per-class performance comparisons show that our improvements are robust and significant.

\begin{table}[]
\centering
\begin{tabular}{|c|c|c|}
\hline
   & Hollywood2 & Olympic Sports\\
   \hline
  L2 Normalization & 57.9\%  & 83.0\% \\
  PN & 66.1\% & 89.4\% \\ 
  RaN & 67.7\% & 92.3\% \\
  Two-level RaN & \textbf{68.6\%} & \textbf{92.5\%} \\
  \hline
\end{tabular}
\caption{\textbf{Performance comparison of different normalization methods on action recognition datasets.} L2 normalization is the baseline method and has also been applied after PN and RaN.  }
\label{tab:rn}
\end{table}

\begin{table}[]
\centering
\begin{tabular}{|c|c|c|}
\hline
  S & Hollywood2 (\% )& Olympic Sports ( \%)\\
   \hline
  1 & $43.8 \pm 24.0$   &  $31.6 \pm 27.5 $\\
  5 & $67.6 \pm 0.2 $& $92.1 \pm 0.9 $  \\ 
  10 & $67.5 \pm 0.3 $& $92.6 \pm 0.4$\\
  50 & $67.6 \pm 0.1 $& $92.7 \pm 0.2 $\\
  100 &  $67.7 \pm 0.1 $& $92.7 \pm 0.1$\\
  \hline
\end{tabular}
\caption{\textbf{Comparison of different subset size S for RaN.} Each experiment is repeated 10 times and the mean values and standard deviations are shown. }
\label{tab:smir}
\end{table}

\subsection{Approximate Ranking}

The exact ranking of RaN requires comparing all the Fisher Vectors in a dataset, which may not be available in many application scenarios such as online learning. Given the high dimensionality of the data, we conjecture that exact ranking may not be necessary. To test our hypothesis, we randomly choose a small subset of Fisher Vectors with size S as seed vectors and compare each Fisher Vector with the seed vectors to get an approximate ranking. S=1 is equivalent to a binarization of the Fisher Vectors. We experiment with different S on Hollywood2 and Olympic Sports datasets and repeat each experiment 10 times.  The results are shown in Table \ref{tab:smir}. Surprisingly, a subset size S as small as 5 is good enough to achieve similar results as the precise ranking. We also observe better performance from Olympic Sports after using approximate ranking. These results show that fine-grain ranking information is not necessary for RaN.

\section{Multi-class Iterative Re-ranking (MIR)}

\begin{figure*}
\centering
    \begin{subfigure}[b]{0.47\textwidth}
                \includegraphics[width=\textwidth]{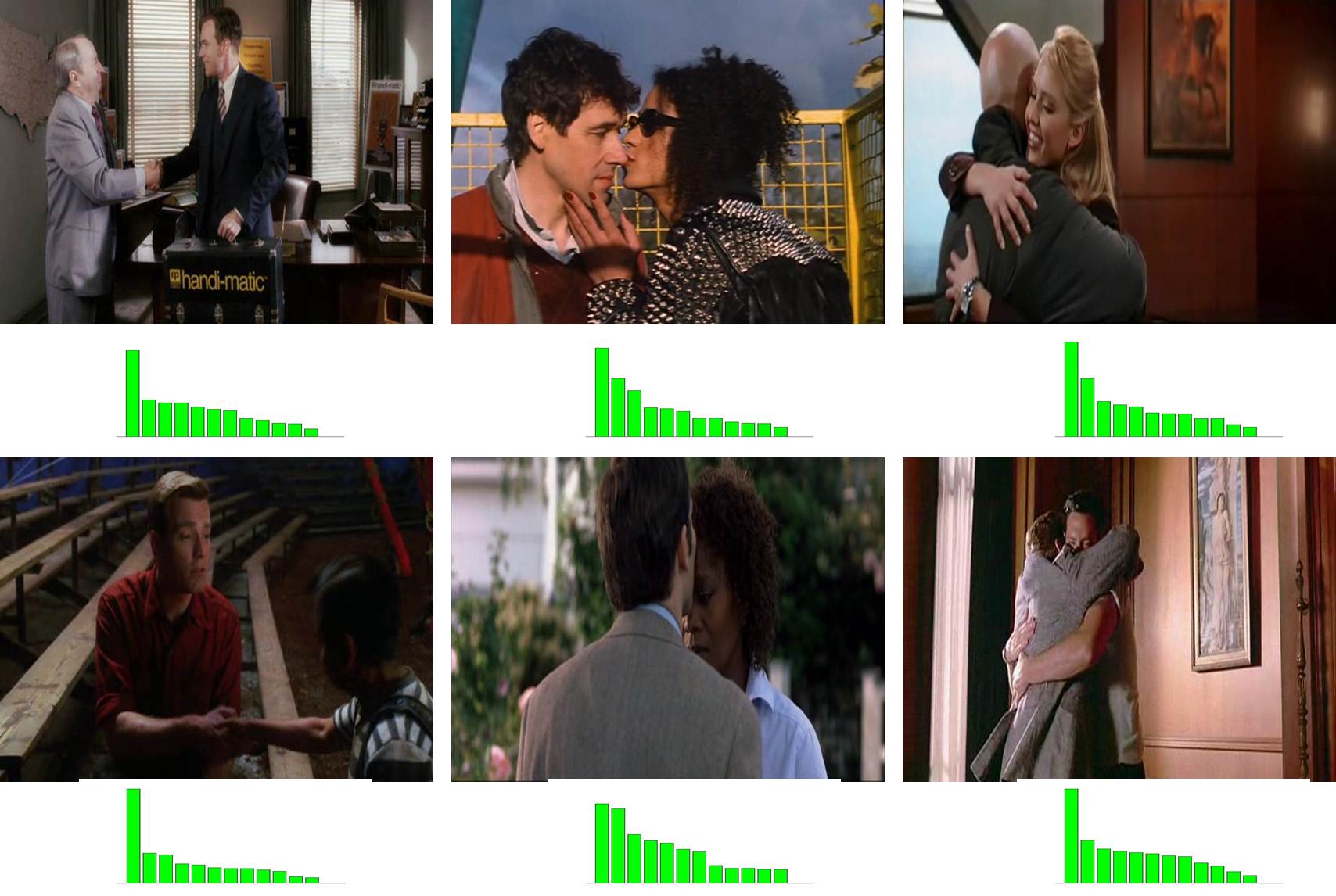}
                    \caption{\textbf{Easy Videos.} The example actions are: HandShake (left), Kiss (middle), HugPerson (right).}
                    \label{fig:ea}
    \end{subfigure}
    \qquad
    \begin{subfigure}[b]{0.47\textwidth}
                \includegraphics[width=\textwidth]{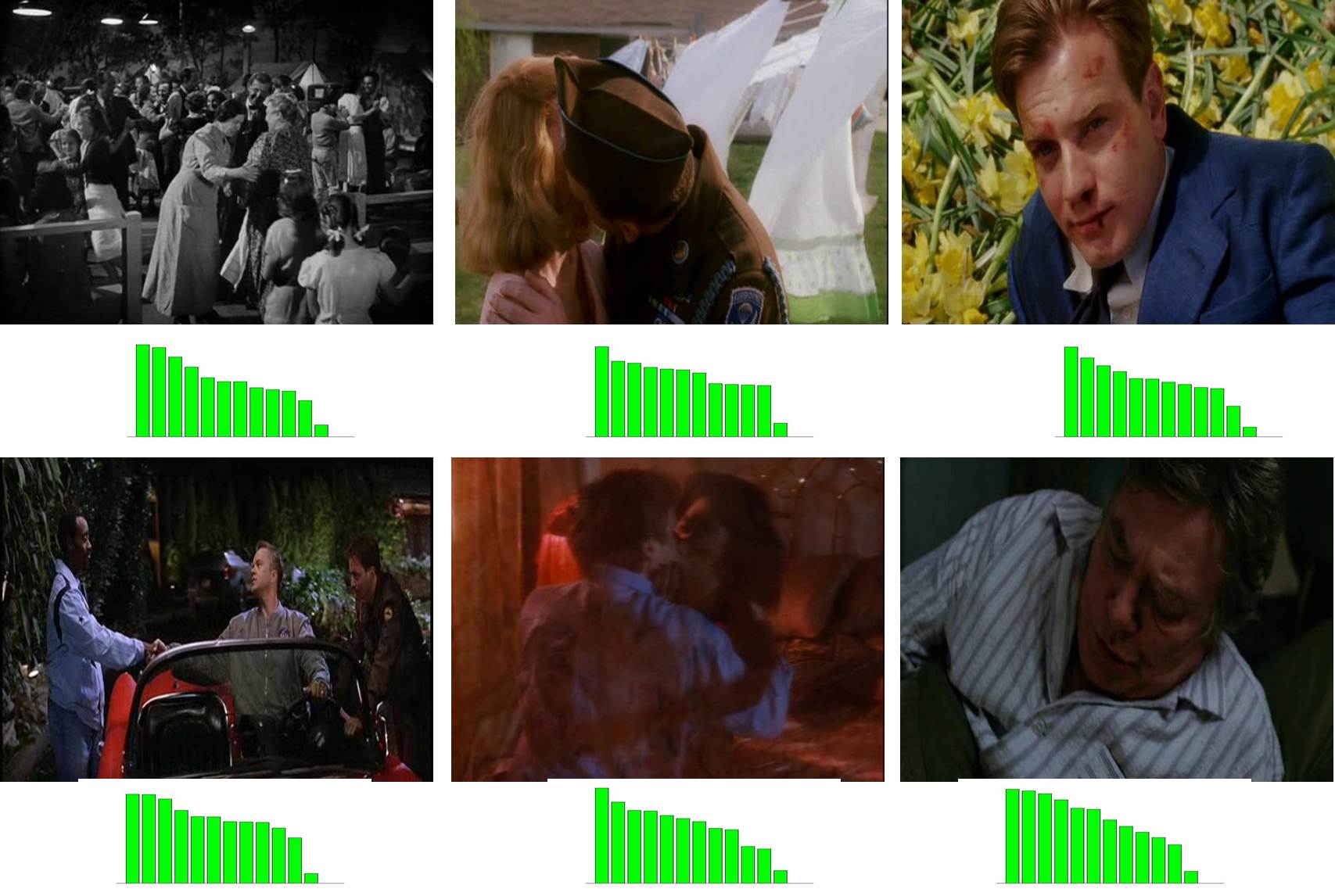}
                    \caption{\textbf{Difficult Videos.} The example actions are: HandShake (left), Kiss (middle), SitUp (right).}
                    \label{fig:eb}
    \end{subfigure}
\caption{\textbf{Illustration of easy and typical videos versus difficult ones.} The bar charts show the sorted predictions of all the classifiers to the example videos. The predictions are normalized so that the values are between 0 and 1. As shown, the predictions of typical easy videos often have one or two dominant classes while those predictions of difficult videos often have much more smooth score distributions. }
\label{fig:examples}
\end{figure*}

\begin{algorithm}
\begin{algorithmic}[1]

\STATE \textbf{Input:} The prediction scores of $K$ class $P\in\mathbb{R}^{N\times K}$;
Re-ranking annealing parameter $\eta$; Re-ranking weighting coefficient
$\beta>0$; Total iteration steps $W$.

\STATE \textbf{Init:}$P^{(0)}=P$

\FOR{\inputencoding{latin1}{$w=1,2,\cdots,W-1$}\inputencoding{latin9}}

\STATE For any instance index $i\in\{1,2,\cdots,N\}$ and class index
$j\in\{1,2,\cdots,K\}$ , 
\begin{align*}
 & \Delta_{i,j}^{(w)}=\mathrm{sort}(\{P_{i,1}^{(w)},P_{i,2}^{(w)},\cdots,P_{i,K}^{(w)}\}\backslash P_{i,j}^{(w)}, \downarrow)\\
 & P_{i,j}^{(w+1)}=P_{i,j}^{(w)}- \eta^{w-1}\sum_{\substack{\substack{r=1}
\\
r\not=j
}
}^{K}\mathrm{e}^{-\beta r}\Delta_{i,j}^{(w)} (r)
\end{align*}

\ENDFOR

\STATE\textbf{Output:} $P^{(W)}$

\end{algorithmic}

\protect\caption{Multi-class Iterative Re-ranking (MIR)}
\label{alg:IRR}
\end{algorithm}

\begin{figure}
\centering
\includegraphics[width=0.45\textwidth]{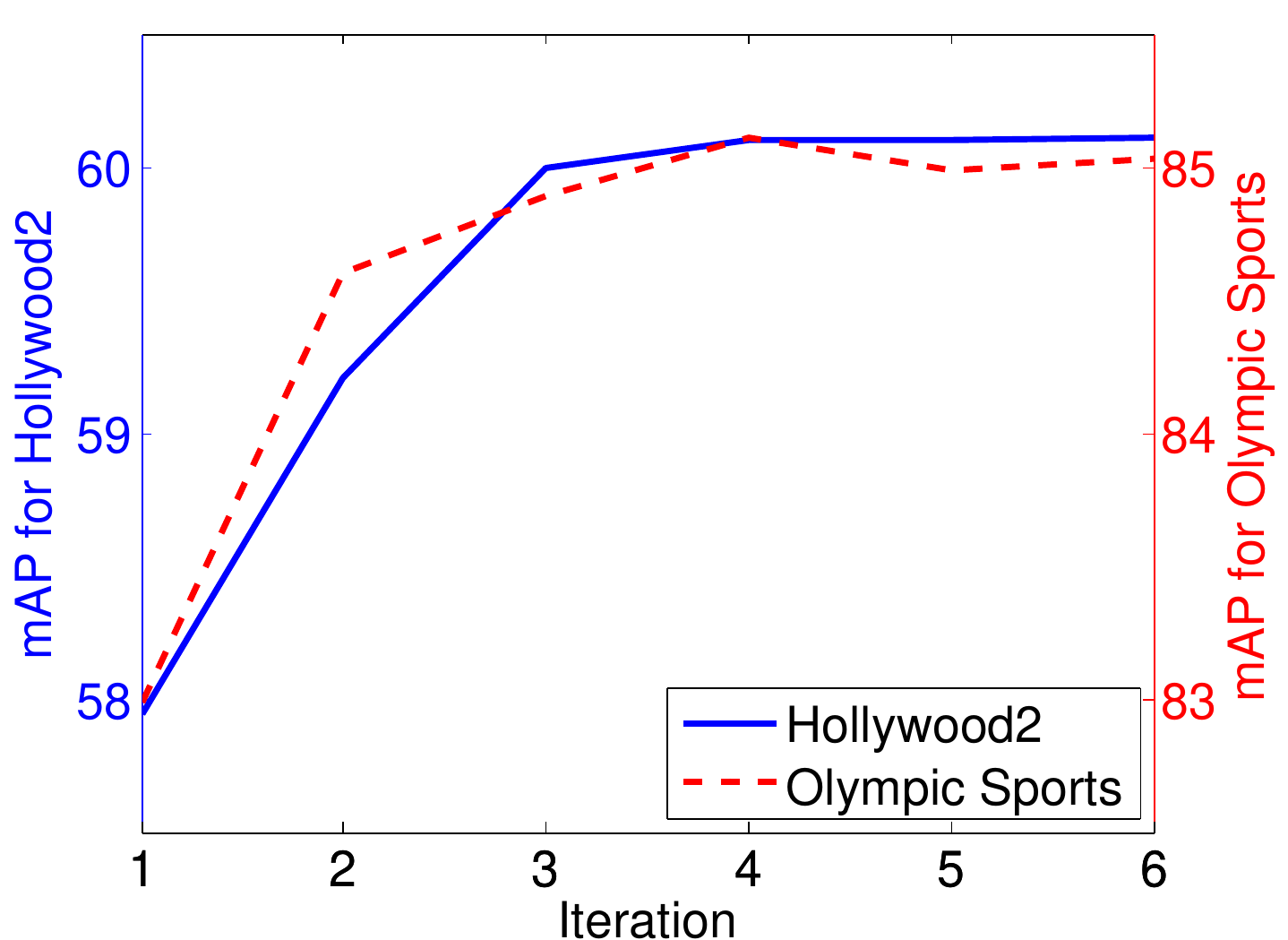}
\caption{ \textbf{MIR performance versus number of iterations.} }
\label{fig:mir}
\end{figure}

\begin{figure}
\centering
    \begin{subfigure}[b]{0.47\textwidth}
                \includegraphics[width=\textwidth]{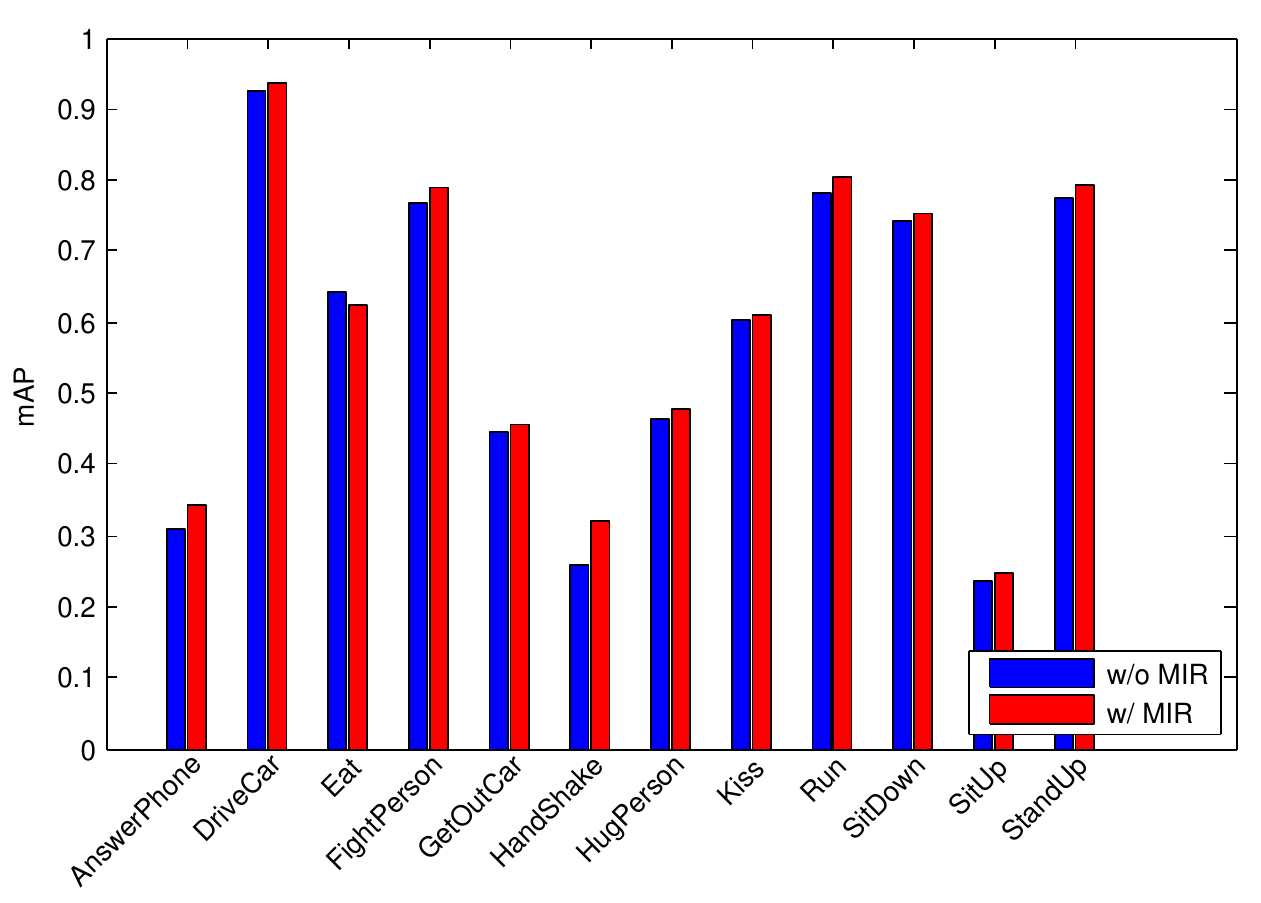}
                    \caption{Hollywood2}
                    \label{fig:ma}
    \end{subfigure}
    \begin{subfigure}[b]{0.47\textwidth}
                \includegraphics[width=\textwidth]{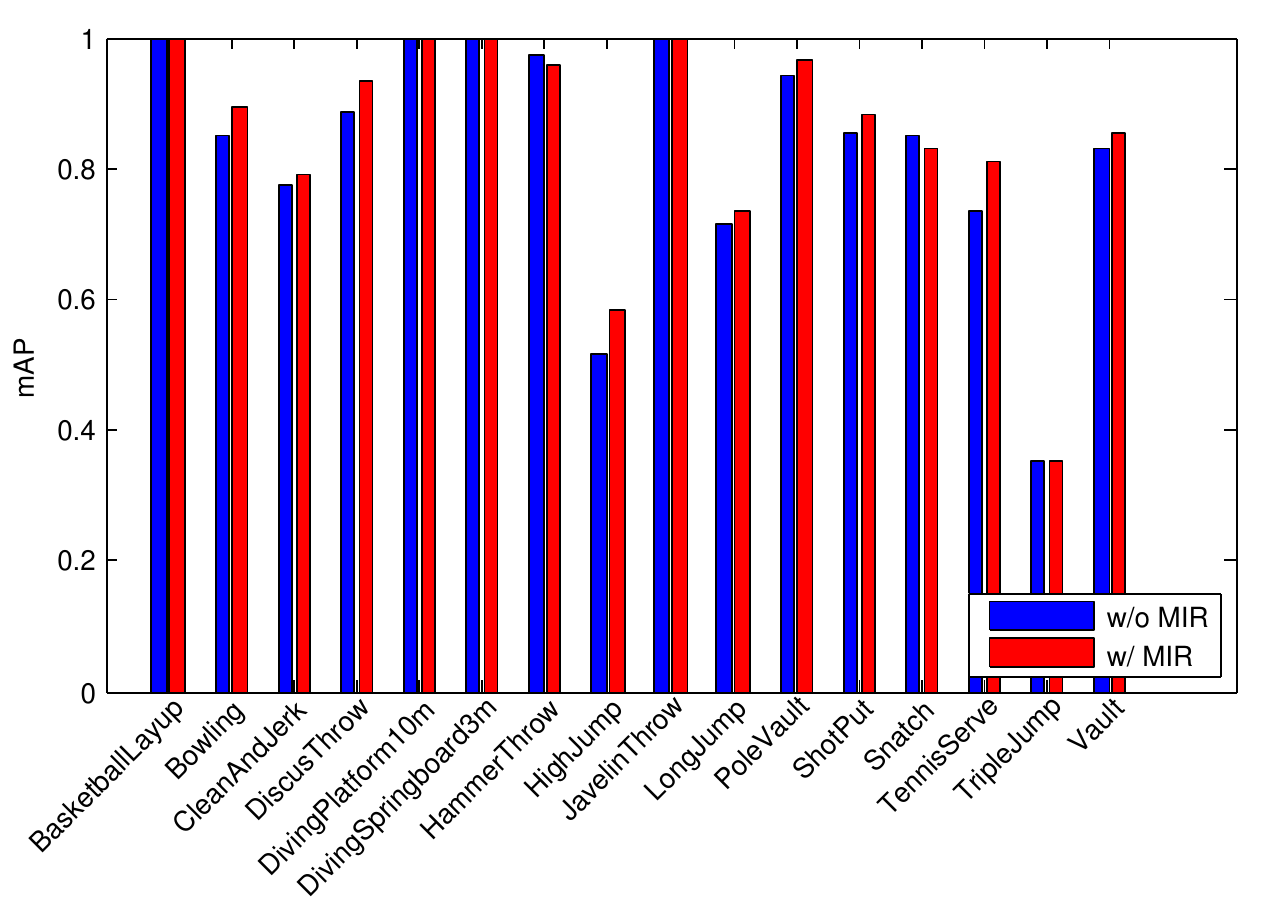}
                    \caption{Olympic Sports}
                    \label{fig:mb}
    \end{subfigure}
\caption{\textbf{Per-class performance comparison of the baseline performances with and without MIR.} }
\label{fig:mn}
\end{figure}


As mentioned previously, traditional curriculum learning \cite{bengio2009curriculum} often relies on human or extra data sources to define the curriculum: easy and typical samples versus difficult ones. In this paper, we instead rely on classifiers of other classes, which are freely available, to define the curriculum. The new way of curriculum definition captures relationship among multiple human activity classes.

As depicted in Figure \ref{fig:examples}, we rank the videos from easy to difficult based on the classifier values. Obviously, easier videos have a much faster roll-off rate of sorted classification predictions. According to this ranking, we discover that videos that have some combinations of the following three scenarios will be more likely to be ranked on the difficult end of the scale: 

\begin{itemize}
\item Contains noisy background motions. If the background of a video contains noisy motions and the target action has a small and weak signal, then the target action would be obscured. For example, the videos of HandShake actions in \ref{fig:eb} are obscured by the background motions and much more difficult to detect than those videos of HandShake actions in \ref{fig:ea}. 
\item Contains multiple actions. If the subject performs multiple actions, the classifiers would be confused as the video features represent a mixture of the multiple actions. For example, the Kiss video in \ref{fig:eb} contain both Kiss and Hug actions (appear in the following frames)  while the Kiss examples in \ref{fig:ea} only contain the Kiss action itself.
\item Ill-defined actions. If the target action is ill-defined by itself, the video would be more difficult to classify. For example, 'SitUp' action is often encapsulated in the 'StandUp' action, which make it harder to classify. 
\end{itemize}

Based on the difficulty scale, we design MIR to re-rank the predictions of classifiers. The intuition behind MIR is that if videos are more difficult to classify, then their predictions are less reliable and their rank should be lowered; if videos are easy and typical, then the predictions should be more reliable and the videos should be ranked higher. The algorithm to re-rank is easy to implement and fast to run. As shown in Algorithm \ref{alg:IRR}, given a score matrix $P\in\mathbb{R}^{N\times K}$ that contains K classifiers' predictions on N videos. We update each score $P_{i,j}$ iteratively by looking at other classifiers' predictions on the same video and reducing the score using the predictions from the other classifiers. The reduction is carried out by first sorting other classifiers' predictions  $\{P_{i,1}^{(w)},P_{i,2}^{(w)},\cdots,P_{i,K}^{(w)}\}\backslash P_{i,j}^{(w)}$ in a descending order and then subtracting the weighted sum of the sorted scores from $P_{i,j}$. We use exponentially decaying weights and the weighting coefficient $\beta$ and the annealing parameter $\eta$ have been set to 1 and 0.5, respectively, throughout the paper. As shown in Figure \ref{fig:mir}, MIR typically converges within 3 or 4 iterations. It improves more than 2\% over the baseline method on both datasets.  We also show the per-class comparison of with and without MIR in Figure \ref{fig:mn}. We can see that for Hollywood2, MIR improves upon the baseline results on 11 out of 12 actions, and for Olympic Sports, MIR improves or gets similar results on 14 out of 16 actions. Different from the improvements of RaN, the improvements of MIR are more evenly distributed among action classes. These per-class performance comparisons again show that the improvements from MIR are robust and significant.

\section{Combined Results: RaN and MIR}
\subsection{Action Recognition}
\begin{figure}
\centering
    \begin{subfigure}[b]{0.47\textwidth}
                \includegraphics[width=\textwidth]{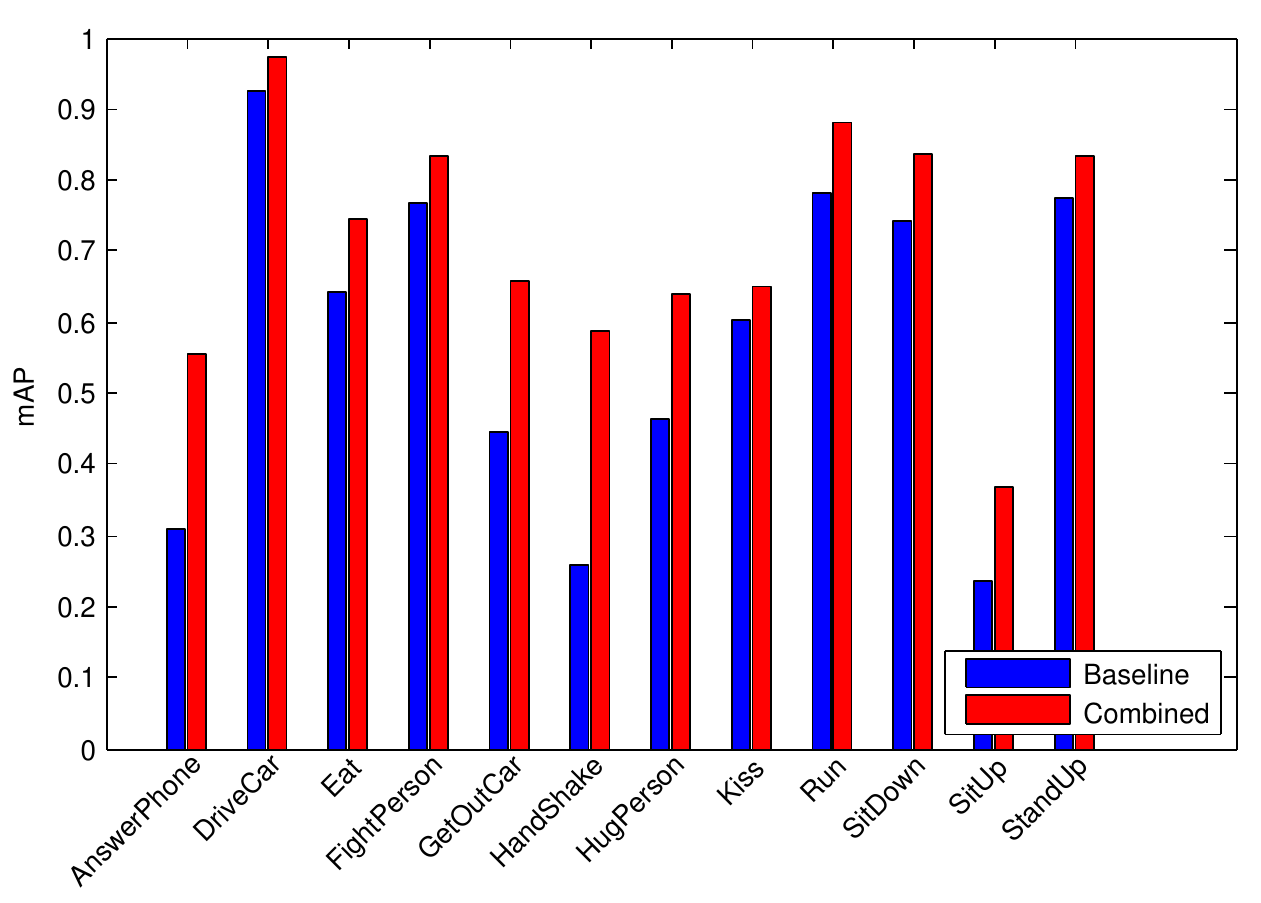}
                    \caption{Hollywood2}
                    \label{fig:aa}
    \end{subfigure}
    \begin{subfigure}[b]{0.47\textwidth}
                \includegraphics[width=\textwidth]{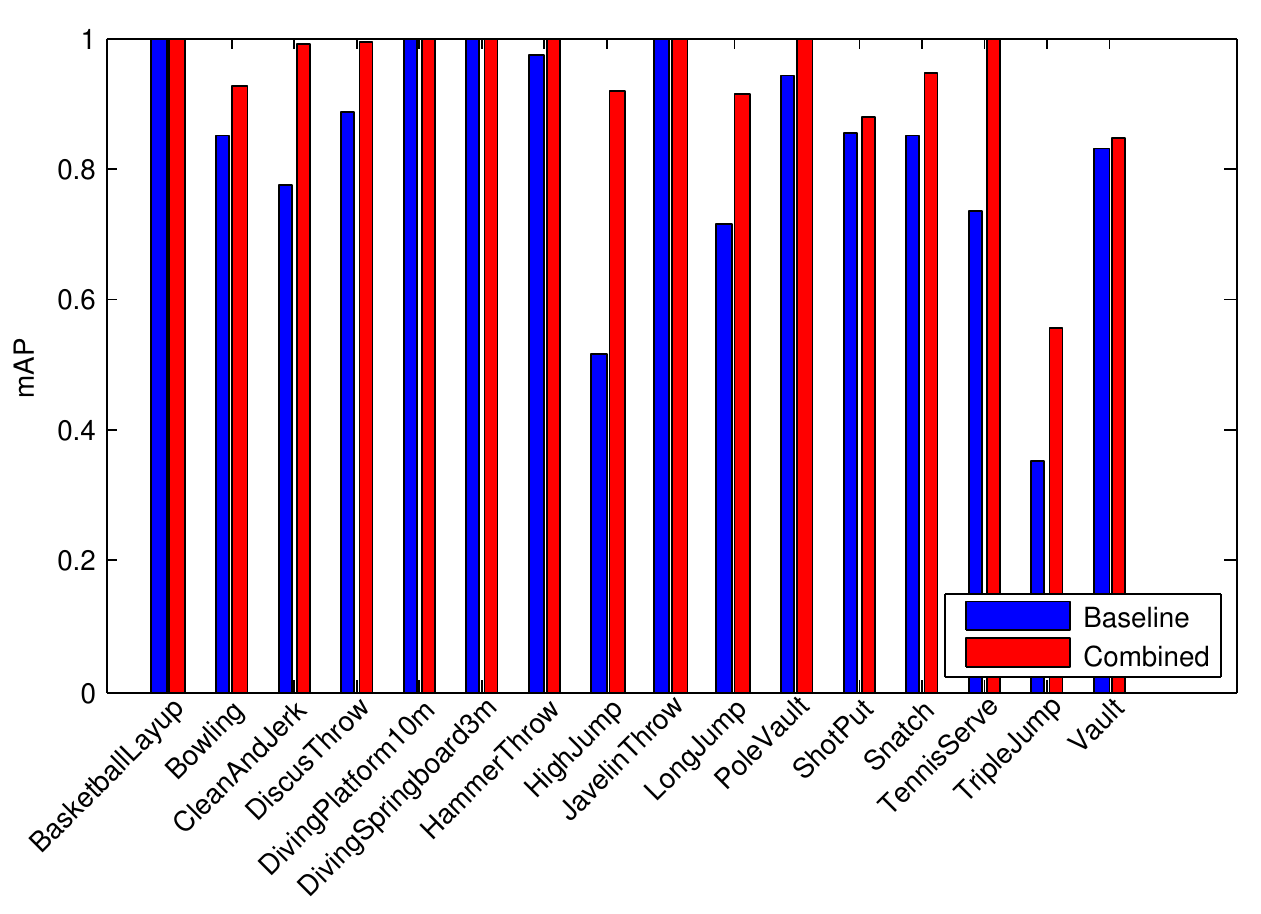}
                    \caption{Olympic Sports}
                    \label{fig:ab}
    \end{subfigure}
\caption{\textbf{Per-class performance comparison of our combined results to the baseline method.} `Combined' indicates applying both RaN and MIR to the baseline method. }
\label{fig:cn}
\end{figure}

If we combine both proposed methods, we observe an even more prominent improvement over the baseline method. We improve upon the baseline method by more than $13\%$ and $10\%$ absolute performance improvement on Hollywood2 and Olympic Sports datasets, respectively. A more detailed comparison is provided in Figure \ref{fig:cn}, from which we can see that our proposed methods together improve the baseline method on all the action classes in Hollywood2. For some of the hard classes like 'Hand Shake' and 'Answer Phone', we can get more than $20\%$ absolute improvement. For Olympic Sports dataset, we observe a similar trend and get 9 out 16 classes with perfect predictions.

In Table \ref{tab:state-of-art}, we compare our combined results to the state-of-the-art performance on Hollywood2 and Olympic Sports datasets. Note that although we list several most recent approaches here for comparison purposes, most of them are not directly comparable to our results due to the use of different features and representations. The most comparable one is Wang \& Schmid \cite{wang2013action}, from which we build our approaches on. Sapienz \textit{et al.} \cite{sapienza2014feature} explored ways to sub-sample and generate vocabularies for Dense Trajectory features. 
Jain et al. \cite{jain2013better}'s approach incorporated a new motion descriptor. Oneata et al.  \cite{oneata2013action} focused on testing Spatial Fisher Vector for multiple actions and event tasks. Adrien \textit{et al.}  \cite{gaidon2014activity} tried to cluster low-level features into mid-level representations in a hierarchical way. The learned hierarchies of mid-level motion components are data-driven decompositions specific to each video. As can be seen, our combined method significantly outperforms these state-of-the-art methods.

\begin{table}
\centering
\begin{tabular}{|l c |l c |}
\hline
\multicolumn{2}{|c|}{Hollywood2 ($\%$)} & \multicolumn{2}{|c|}{Olympics Sports ( $\%$)}\\ \hline
Sapienz \textit{et al.} \cite{sapienza2014feature}  & 59.6  & Jain \textit{et al.} \cite{jain2013better}  & 83.2\\
Jain \textit{et al.} \cite{jain2013better}    & 62.5  & Oneata \textit{et al.} \cite{oneata2013action}  & 84.6\\\
Oneata \textit{et al.} \cite{oneata2013action} &  63.3 &   Adrien \textit{et al.} \cite{gaidon2014activity}  & 85.5 \\
Wang \textit{et al.} \cite{wang2013action}  &64.3 &  Wang \textit{et al.} \cite{wang2013action}  & 91.1\\
Lan \textit{et al.} \cite{lan2014beyond}   & 68.0 & Lan \textit{et al.} \cite{lan2014beyond}   & 91.4\\
\hline 
Combined  & \textbf{71.4} & Combined &  \textbf{93.6}\\ 
\hline
\end{tabular}
\caption{\label{tab:state-of-art}\textbf{Comparison of our results to the state-of-the-arts on the action recognition task.} `Combined' indicates applying both RaN and MIR to the baseline method. }
\end{table}

\subsection{Multimedia Event Detection}
\paragraph{Problem formulation}
Given a collection of videos, the goal of an event detection task is to detect events of interest such as birthday party and parade, solely based on the video content. The task is challenging due to complex actions and scenes. By evaluating on this task, we examine how RaN and MIR behavior on difficult event detection tasks. 

\paragraph{Benchmark datasets}
TREC Video Retrieval Evaluation (TRECVID) Multimedia Event Detection (MED) \cite{over2013trecvid} is a task organized by NIST (National Institute of Standards and Technology) aimed at encouraging new technologies for detecting complex events such as \textit{having a birthday party} and \textit{rock climbing}. Until 2014, NIST has built up a database that contains 8000 hours of videos and 40 events, which is by far the largest human labeled event detection collection. MEDTEST13, 14 are two standard evaluation datasets released by NIST in 2013 and 2014, respectively. Each of them contains around 10 percent of the whole MED collection and has 20 events and consist of two tasks, i.e. EK100 and EK10. EK100 has 100 positive training samples per event while EK10 only has 10. Together, each dataset has 8000 training samples and 24000 testing samples. 

\paragraph{Experimental settings} This is similar to the settings discussed in section \ref{action_recognition}. 

\paragraph{Results} Table \ref{tab:med} lists the overall mAP on all four datasets. The baseline method is a conventional IDT representation. First, on all four datasets, RaN notably improves the performance of the conventional IDT representation. These results demonstrate that RaN is robust across tasks with different difficulty levels. For MIR, in EK10 scenario where the baseline performance is unusually low, MIR hurts the performance due to the inaccurate curriculum estimation; in EK100 settings that have comparatively reasonable baseline performance, MIR manages to achieve noticeable improvements, though not as much as the improvements in action recognition tasks where the baseline performances are much higher. These results reveal the fact that for MIR being useful, the baseline performance should be reasonably accurate. Finally, the combined results, though not always the best, improve upon the baseline method by around $4\%$ on EK100 tasks and about $3\%$ on EK10 tasks. It is worth emphasizing that MED is such a challenging task that $3\%$ of absolute performance improvement is significant.

\begin{table}
\centering
\begin{tabular}{|c | c c|c c|}
\hline
 & \multicolumn{2}{|c|}{MEDTEST13}  & \multicolumn{2}{|c|}{MEDTEST14}\\
\hline 
 & EK10& EK100  & EK10 & EK100 \\\hline
  Baseline & 17.0 & 33.6& 12.0  & 26.2  \\ 
  RaN & \textbf{20.2} & 36.6 & \textbf{15.4} & 29.3\\
  MIR & 16.7 & 34.2 & 11.3 & 26.6 \\
  Combined & 20.0 & \textbf{37.5} &  14.9 & \textbf{29.9} \\ 
\hline
\end{tabular}
\caption{\label{tab:med}\textbf{Performance Comparison on the MED task.}}
\end{table}

\section{Conclusions and Discussions}
This paper has introduced two ranking-based methods to improve the performance of state-of-the-art action recognition systems. RaN ranks each dimension of Fisher Vectors and uses the ranks to replace the original vectors. It improves the classic PN method by addressing the sparse and bursty distribution problem of Fisher Vectors and VLADs. MIR iteratively uses the predictions of other classifiers to rank videos from easy to difficult task levels and re-ranks the predictions accordingly. These two methods together significantly improve the performance of Fisher Vectors on six real-world datasets and set new state-of-the-art results for two benchmark action datasets including Hollywood2 and Olympic Sports. Future work will be investigating how to use RaN for local descriptors. In addition, we would like to apply the proposed methods to VLAD and other video features such as deep neural network features. 

{\small
\bibliographystyle{ieee}
\bibliography{egbib}
}

\end{document}